\documentclass[runningheads]{llncs}
\usepackage{graphicx}
\usepackage{comment}
\usepackage{amsmath,amssymb} 
\usepackage{color}

\usepackage{stfloats}
\usepackage{epsfig}
\usepackage{caption}
\usepackage{booktabs}
\usepackage{multirow}
\usepackage{amssymb}
\usepackage{pifont}
\newcommand{\cmark}{\ding{51}}%
\newcommand{\xmark}{\ding{55}}%
\newcommand{\bc}[1]{\textcolor{blue}{#1}}
\newcommand{\rc}[1]{\textcolor{red}{#1}}
\newcommand{\gc}[1]{\textcolor{green}{#1}}

\usepackage{wrapfig}
\begin{document}
\pagestyle{headings}
\mainmatter
\def\ECCVSubNumber{5897}  

\title{Guidance and Evaluation: Semantic-Aware Image Inpainting for Mixed Scenes} 

\titlerunning{Semantic-Aware Image Inpainting for Mixed Scenes}
%
\author{Liang Liao\inst{1,2}\orcidID{0000-0002-2238-2420} \and Jing Xiao\inst{1,2}\thanks{Corresponding author: Jing Xiao}\orcidID{0000-0002-0833-5679} \and
Zheng Wang\inst{2}\orcidID{0000-0003-3846-9157} \and Chia-Wen Lin\inst{3}\orcidID{0000-0002-9097-2318} \and \\ Shin'ichi Satoh\inst{2}}
\authorrunning{L. Liao, J. Xiao, Z. Wang, C. Lin, S. Satoh}
%
\institute{National Engineering Research Center for Multimedia Software, School of Computer Science, Wuhan University \and National Institute of Informatics \and Department of Electrical Engineering, National Tsing Hua University \\
\email{\{liang, wangz, satoh\}@nii.ac.jp; jing@whu.edu.cn; cwlin@ee.nthu.edu.tw}}
\maketitle

\begin{abstract}
Completing a corrupted image with correct structures and reasonable textures for a mixed scene remains an elusive challenge. Since the missing hole in a mixed scene of a corrupted image often contains various semantic information, conventional two-stage approaches utilizing structural information often lead to the problem of unreliable structural prediction and ambiguous image texture generation. In this paper, we propose a Semantic Guidance and Evaluation Network (SGE-Net) to iteratively update the structural priors and the inpainted image in an interplay framework of semantics extraction and image inpainting. It utilizes semantic segmentation map as guidance in each scale of inpainting, under which location-dependent inferences are re-evaluated, and, accordingly, poorly-inferred regions are refined in subsequent scales. Extensive experiments on real-world images of mixed scenes demonstrated the superiority of our proposed method over state-of-the-art approaches, in terms of clear boundaries and photo-realistic textures.
\end{abstract}

\section{Introduction}

Image inpainting refers to the task of filling the missing area in a scene with synthesized content. Due to its wide applications in photo editing, de-caption, damaged image repairing, error concealment in data transmission, etc., it has drawn great attention in the field of computer vision and graphics \cite{barnes2009patchmatch,bertalmio2000image,criminisi2004region,hays2007scene,sun2005image}. Recent learning-based methods have achieved great success in filling large missing regions with plausible contents of various simple scenes \cite{pathak2016context,wang2018image,xiao2019cisi,yu2018generative,yu2019free,zhang2019base,8067496,zheng2019pluralistic}. However, these existing methods still encounter difficulties while completing images of a mixed scene, that composes of multiple objects with different semantics.  

\begin{figure}[t]
\centering
\begin{tabular}{cccccc}
			\includegraphics[width=0.16\linewidth]{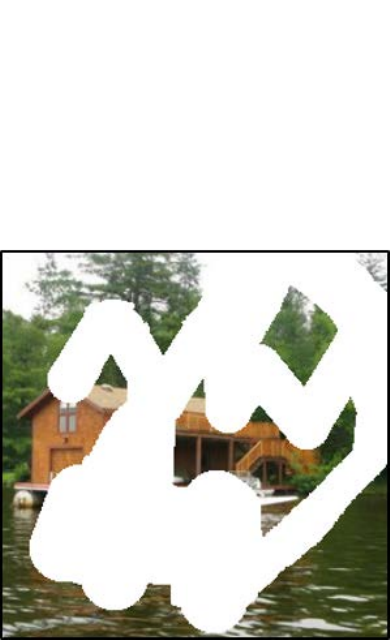} &
			\includegraphics[width=0.16\linewidth]{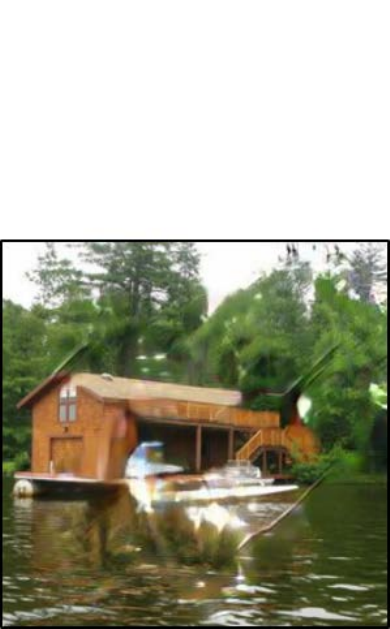} &
			\includegraphics[width=0.16\linewidth]{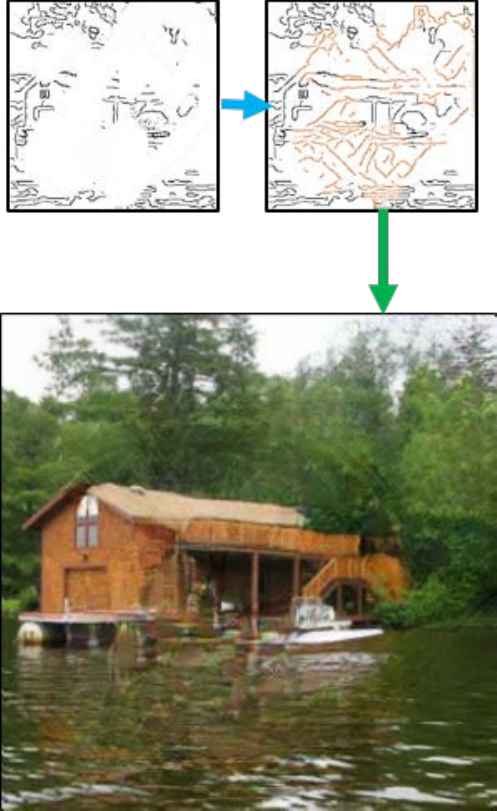} &
			\includegraphics[width=0.16\linewidth]{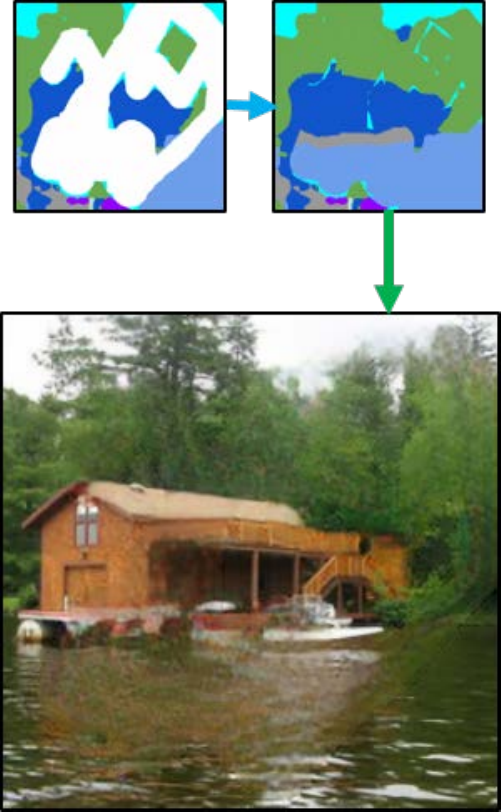} &
			\includegraphics[width=0.163\linewidth]{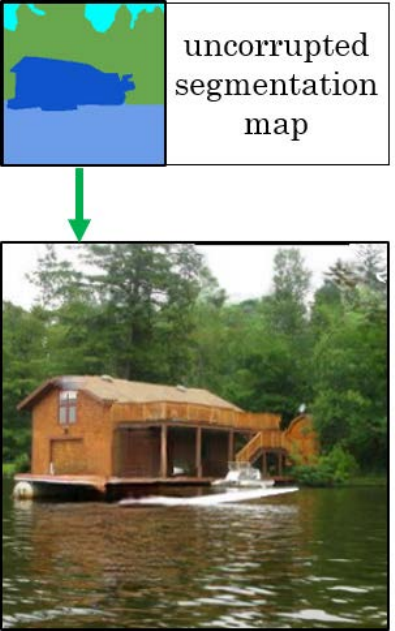} &
			\includegraphics[width=0.16\linewidth]{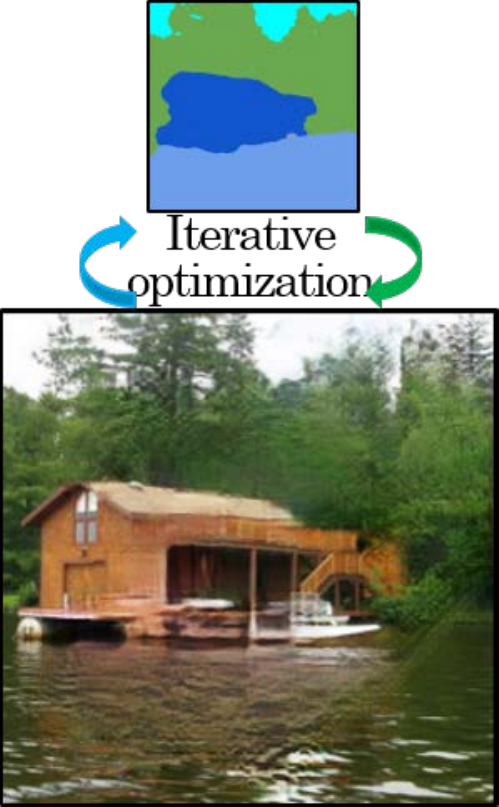} \\
			\tiny{(a) Input} & \tiny{(b) GC} & \tiny{(c) EC} & \tiny{(d) SPG} & \tiny{(e) Ideal case} & \tiny{(f) SGE-Net (ours)}\\
	\end{tabular}
	\linespread{1}
\caption{\scriptsize{Comparison of the inpainting results for a mixed scene: (b) GC~\cite{yu2019free} without structural information; (c) EC~\cite{nazeri2019edgeconnect} with predicted edges; (d) SPG~\cite{song2018spg} with less reliable predicted semantic segmentation; (e) semantic-guided inpainting with an uncorrupted segmentation map; and (f) the proposed SGE-Net with iteratively optimized semantic segmentation. [Best viewed in color]}}
\label{fig:intro_motivation}
\end{figure}

Existing learning-based image inpainting methods typically fill missing regions by inferring the context of corrupted images~\cite{iizuka2017globally,pathak2016context,xie2019image,yu2018generative,yu2019free}. However, in a mixed scene, the prior distributions of various semantics are different and various semantic regions also contribute differently to pixels in the missing regions, thus uniformly mapping different semantics onto a single manifold in the context-based methods often leads to unrealistic semantic content as illustrated in Fig.~\ref{fig:intro_motivation}(b).

To address this issue, low to mid-level structural information \cite{liao2018edge,song2018spg,xiong2019foreground,zhao2019parallel} was introduced to assist image inpainting. These methods extract and reconstruct the edges or contours in the first stage and complete an image with the predicted structural information in the second stage. The spatial separation by the structures helps to alleviate the blurry boundary problem. These methods, however, ignore the modeling of semantic content, which may result in ambiguous textures at the semantic boundaries. Moreover, the performance of the two-stage inpainting process highly relies on the reconstructed structures from the first stage, but the unreliability of the edge or contour connections largely increases in a mixed scene (Fig.~\ref{fig:intro_motivation}(c)). As revealed in \cite{xiao2019cisi} that human beings perceive and reconstruct the structures under the semantic understanding of a corrupted image, it is natural to involve semantic information in the process of image inpainting. 

In this paper, we show how semantic segmentation can effectively assist image inpainting of a mixed scene based on two main discoveries: {\bf semantic guidance and segmentation confidence evaluation}. Specifically, a semantic segmentation map carries pixel-wise semantic information, providing the layout of a scene as well as the category, location and shape of each object. It can assist the learning of different texture distributions of various semantic regions. Moreover, the intermediate confidence score derived from the segmentation process can offer a self-evaluation for an inpainted region, under the assumption that ambiguous semantic contents usually cannot lead to solid semantic segmentation results. 

To the best of our knowledge, a similar work making use of semantic segmentation information for image inpainting is SPG proposed in \cite{song2018spg}, which is also a two-stage process. It extracts and reconstructs a segmentation map, and then utilizes the map to guide image inpainting. Thanks to the helpful semantic information carried in the segmentation map, SPG can effectively improve inpainting performance compared to those methods without a semantic segmentation map. Nevertheless, it is hard to predict reliable semantics about a region when its context information is largely missing, especially in mixed scene. As a result, its performance can be significantly degraded by such unreliable semantic region boundaries and labels predicted by the semantic segmentation. Such performance degradation is evidenced in Fig.~\ref{fig:intro_motivation}(d), from which we can observe blurry and incorrect inpainted textures generated by SPG. By contrast, segmentation-guided inpainting can achieve high-quality image completion provided that a reliable segmentation map (i.e., the segmentation map of uncorrupted image) is given as illustrated in Fig.~\ref{fig:intro_motivation}(e). Therefore, to make the best use of semantic information carried in the segmentation map for image inpainting, how to predict a reliable semantic segmentation map, even if part of an image is corrupted, is the key.

To address the above problems, we advocate that the interplay between the two tasks, semantic segmentation and image inpainting, can effectively improve the reliability of the semantic segmentation map from a corrupted image,  which will in turn improve the performance of inpainting as illustrated in Fig.~\ref{fig:intro_motivation}(f). To this end, we propose a novel \textbf{S}emantic \textbf{G}uidance and \textbf{E}valuation \textbf{Net}work (\textbf{SGE-Net}) that makes use of the interplay between semantic segmentation and image inpainting in a coarse-to-fine manner. Experiments conducted on the datasets containing mixtures of multiple semantic regions demonstrated the effectiveness of our method in completing a corrupted mixed scene with significantly improved semantic contents.

Our contributions are summarized as follows:

1) We show that the interplay between semantic segmentation and image inpainting in a coarse-to-fine manner can effectively improve the performance of image inpainting by simultaneously generating an accurate semantic guidance from merely an input corrupted image.

2) We are the first to propose a self-evaluation mechanism  for image inpainting through segmentation confidence scoring to effectively localize the predicted pixels with ambiguous semantic meanings, which enables the inpainting process to update both contexts and textures progressively.

3) Our model outperforms the state-of-the-art methods, especially on mixed scenes with multiple semantics, in the sense of generating semantically realistic contexts and visually pleasing textures.

\section{Related Work}

\subsection{Deep Learning-Based Inpainting}

Deep learning-based image inpainting approaches \cite{li2017generative,pathak2016context,yeh2017semantic} are generally based on generative adversarial networks (GANs) \cite{goodfellow2014generative,radford2015unsupervised,wang2018cascaded} to generate the pixels of a missing region. For instance, Pathak \textit{et al.} introduced Context Encoders \cite{pathak2016context}, which was among the first approaches in this kind. The model was trained to predict the context of a missing region but usually generates blurry results. Based on the Context Encoders model, several methods were proposed to better recover texture details through the use of well-designed loss functions \cite{dosovitskiy2016generating,iizuka2017globally,li2017generative}, neural patch synthesis \cite{yang2017high}, residual learning \cite{demir2017deep,yi2019multi}, feature patch matching \cite{song2018contextual,yan2018shift,yu2018generative,zeng2019learning}, content and style disentanglement \cite{gilbert2018disentangling,wang2019learning,xiao2019cisi}, and others \cite{ma2019coarse,wang2019musical,wang2018image}. Semantic attention was further proposed to refine the textures in \cite{liu2019coherent}. However, most of the above methods were designed for dealing with rectangular holes, but cannot effectively handle large irregular holes.
To fill irregular holes, Liu \textit{et al.} \cite{liu2018image} proposed a partial convolutional layer, which calculates a new feature map and updates the mask at each layer. Later, Yu \textit{et al.} \cite{yu2019free} proposed a gated convolutional layer based on the models in \cite{yu2018generative} for  irregular image inpainting. While these methods work reasonably well for one category of objects or background, they can easily fail if the missing region contains multiple categories of scenes.

\subsection{Structural Information-Guided Inpainting}
Recently, structural information was introduced in learning-based framework to assist the image inpainting process. These methods are mostly based on two-stage networks, where missing structures are reconstructed in the first stage and then used to guide the texture generation in the second stage. Edge maps were first introduced by Liao \textit{et al.} \cite{liao2018edge} as a structural guide to the inpainting network. This idea is further improved by Nazeri \textit{et al.} \cite{nazeri2019edgeconnect} and Li \textit{et al.} \cite{li2019progressive} in terms of better edge prediction. Similar to edge information, object contours were used by Xiong \textit{et al.} \cite{xiong2019foreground} to separately reconstruct the foreground and background areas. Ren \textit{et al.} \cite{ren2019structureflow}  proposed using smoothed images to carry additional image information other than edges as prior information. Considering semantic information for the modeling of texture distributions, SPG proposed in \cite{song2018spg} predicts the semantic segmentation map of a missing region as a structural guide. The above-mentioned methods show that the structure priors effectively help improve the quality of the final completed image. However, how to reconstruct correct structures remains challenging, especially when the missing region becomes complex.

\section{Approach}

As illustrated in Figs.~\ref{fig:intro_motivation}(d)-(f), the success of semantic segmentation-guided inpainting depends on a reliable segmentation map, which is hard to obtain from an image with a corrupted mixed scene. To address this issue, we propose a novel method to progressively predict a reliable segmentation map from a corrupted image through the interplay between semantic segmentation and image inpainting in a coarse-to-fine manner. To verify how semantic information boosts image inpainting, two networks are proposed. As a baseline, the first one uses only semantic guidance on image inpainting. Moreover, the semantic evaluation is added in the second network as an advanced strategy.

We first introduce some notations used throughout this paper. Given a corrupted image \(X\) with a binary mask \(M\) (0 for holes), and the corresponding ground-truth image  ${Y}$, the inpainting task is to generate an inpainted image $\hat{Y}$ from  $X$ and $M$. Given a basic encoder-decoder architecture of $L$ layers, we denote the feature maps from deep to shallow in the encoder as $\phi^L$, $\phi^{L-1}$, ..., $\phi^l$, ...,  $\phi^1$, and in the decoder as $\varphi^L$,  $\varphi^{L-1}$..., $\varphi^l$, ...,  $\varphi^1$.

\begin{figure}[t]
\begin{center}
   \includegraphics[width=\linewidth]{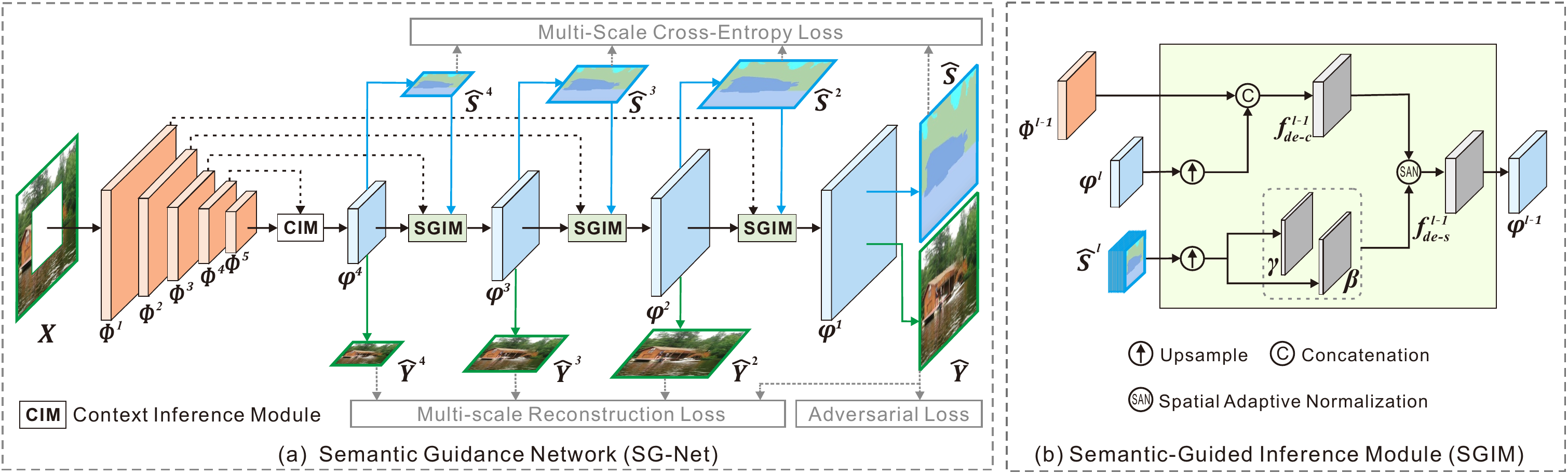}
\end{center}
   \caption{\scriptsize{Proposed Baseline: Semantic Guidance Network (SG-Net). It iteratively updates the contextual features in a coarse-to-fine manner. SGIM updates the predicted context features based on the segmentation map at the next scale.}}
\label{fig:frame_simple}
\end{figure}

\subsection{Semantic Guidance Network (SG-Net)}

The SG-Net architecture is shown in Fig.~\ref{fig:frame_simple}(a). The encoder is used to extract the contextual features of a corrupted image. The decoder then updates the contextual features to predict the semantic segmentation maps and inpainted images simultaneously in a multi-scale manner. Based on this structure, semantic guidance takes effect in two aspects. First, the semantic supervisions are added to guide the learning of contextual features at different scales of the decoder. Second, the predicted segmentation maps are involved in the inference modules to guide the update of the contextual features at the next scale. Being different from the two-stage process\cite{liao2018edge,nazeri2019edgeconnect,song2018spg}, the supervision of semantic segmentation on the contextual features enables them to carry the semantic information, that helps the decoder learn better texture models for different semantics. 

The corrupted image is initially completed in the feature level through a Context Inference Module (CIM). After that, the image inpainting and semantic segmentation interplay with each other and are progressively updated across scales. Two branches are extended from the contextual features at each scale of the decoder to generate multi-scale completed images  $\hat{Y}^{L-1}$, ..., $\hat{Y}^{l}$, ..., $\hat{Y}^{1}$ and their semantic segmentation maps $\hat{S}^{L-1}$, ..., $\hat{S}^{l}$, ..., $\hat{S}^{1}$.

\begin{equation}
 \centering
\hat{Y}^{l} = h(\varphi^l),\quad \hat{S}^{l} = g(\varphi^l),
\end{equation}
where $h(\cdot)$ and $g(\cdot)$ denote the inpainting branch and segmentation branch, respectively. 

\textbf{Semantic-Guided Inference Module (SGIM)} SGIM is designed to make an inference and update the contextual features at the next scale $\varphi^{l-1}$. As shown in Fig.~\ref{fig:frame_simple}(b), SGIM takes three types of inputs: two of them are the current contextual features $\varphi^{l}$ and the skip features of the next scale $\phi^{l-1}$ from the encoder. The third input is the segmentation map $\hat{S}^{l}$, which is used to formalize the textures under the assumption that those regions of the same semantic class should have similar textures. The inference process can be formulated as follows:
\begin{equation}
 \centering
\varphi^{l-1} = infer(\varphi^{l}, \phi^{l-1}, \hat{S}^l),
 \label{eq:infer}
\end{equation}
where $infer(\cdot)$ is the process of updating the contextual features in SGIM.

\begin{figure}[tb]
\begin{center}
   \includegraphics[width=\linewidth]{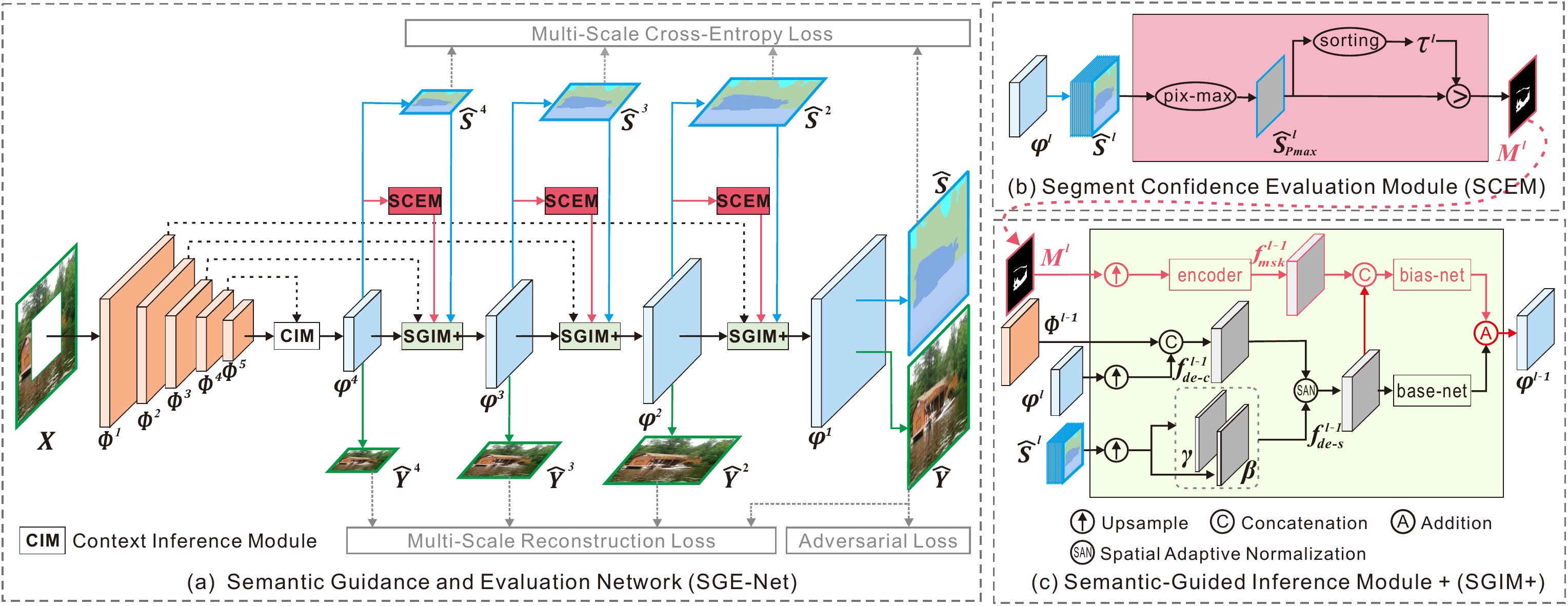}
\end{center}
   \caption{\scriptsize{Proposed Semantic Guidance and Evaluation Network (SGE-Net). It iteratively evaluates and updates the contextual features through the SCEM and SGIM+ modules in a coarse-to-fine manner, where SCEM identifies the pixels where the context needs to be corrected, while SGIM+ updates the predicted context features representing the incorrect pixels located by SCEM.}}
\label{fig:frame_full}
\end{figure}

To update the contextual features based on segmentation map $\hat{S}^l$, we follow the image generation approach in \cite{park2019semantic}, which adopts spatial adaptive normalization to propagate semantic information into the predicted images for achieving effective semantic guidance. The contextual features $f_{de-s}^{l-1}$ are updated as follows:
\begin{equation}
 \centering
f_{de-s}^{l-1}=\gamma\odot\frac{f_{de-c}^{l-1}-\mu}{\sigma}+\beta,
 \label{eq:feature_updating}
\end{equation}
where $(\gamma, \beta)$ is a pair of affine transformation parameters modeled from segmentation map $\hat{S}^l$, $\mu$ and $\sigma$ are the mean and standard deviation of each channel in the concatenated feature vector $f_{de-c}^{l-1}$ generated from $\phi^l$ and $\varphi^{l-1}$. \(\odot\) denotes element-wise multiplication.

\subsection{Semantic Guidance and Evaluation Network (SGE-Net)}
To deeply exploit how segmentation confidence evaluation can help correct the wrongly predicted pixels, we add the Segmentation Confidence Evaluation Module (SCEM) on each decoder layer of SG-Net. The evaluation is performed under the assumption that predicted ambiguous semantic content would result in low confidence scores during the semantic segmentation process. Therefore, we introduce the \textbf{segmentation confidence scoring} after each decoding layer to self-evaluate the predicted region. The reliability mask is then feed to the next scale, which can be used to identify those to-be-updated contextual features that contribute to the unreliable area. This module enables the proposed method to correct the mistakes in those regions completed at the previous coarser scale. Fig.~\ref{fig:frame_full}(a) illustrates the detailed architecture of SGE-Net.

\textbf{Segmentation Confidence Evaluation Module (SCEM)} The output of the semantic segmentation branch is a $k$-channel probability map. The confidence score at every channel of a pixel in the map signifies how the pixel looks like a specific class. Based on the scores, we assume that an inpainted pixel is unreliable if it has low scores for all semantic classes. 

The framework of SCEM is depicted in Fig.~\ref{fig:frame_full}(b). Taking the segmentation probability map at a certain scale $\hat{S}^{l}$, we generate a reliability mask $M^{l}$ to locate those pixels which might have unreal semantic meaning. We first generate a max-possibility map $\hat{S}_{P_{max}}^{l}$ by assigning each pixel with the highest confidence score of $k$ channels in $\hat{S}^{l}$. Then, the mask value of pixel $(x,y)$ in the reliability mask is decided by judging whether the max-confidential score at each pixel location exceeds a threshold $\tau^l$.

\begin{equation}
 \centering
M^{l}(x,y)=\begin{cases}
1,& \hat{S}_{P_{max}}^{l}>\tau^l \\
0,& \text{otherwise} 
\end{cases},
 \label{eq:seg_mask}
\end{equation}
where $\tau^l$ is decided by the percentile of the sorted confidence value.

\textbf{Enhanced SGIM (SGIM+)}  In order to correct the pixels marked as unreliable from the SCEM, SGIM+ takes the reliability mask $M^{l}$ as the fourth input to update the current context features (as shown in Fig.~\ref{fig:frame_full}(c)). The formulation of the inference process can be updated as follows:
\begin{equation}
 \centering
\varphi^{l-1} = infer(\varphi^{l}, \phi^{l-1}, \hat{S}^l, M^l).
 \label{eq:infer_updating}
\end{equation}

To enable the dynamic corrections of semantics, we
introduced a bias-net $F_{bi}^l$ in correspondence to the original network branch between the feature $f_{de-s}^{l-1}$ and $\varphi^{l-1}$ in the previous version of SGIM, which we call base-net $F_{ba}^l$. The new reliability mask $M^l$ is fed into the bias-net to learn residuals to rectify the basic contextual features from the base-net. The new contextual features at the next scale can be formulated as
\begin{equation}
 \centering
\varphi^{l-1} = F_{ba}^l(f_{de-s}^{l-1})+F_{bi}^l(f_{de-s}^{l-1}\oplus F(M^l)),
 \label{eq:context}
\end{equation}
where $\oplus$ denotes the concatenation operation and $F$ represents the convolutions to translate the reliability mask into feature maps.

\subsection{Training Loss Function}

The loss functions comprise loss terms for both image inpainting and semantic segmentation. For image inpainting, we adopt the multi-scale reconstruction loss to refine a completed image and the adversarial loss to generate visually realistic textures. For semantic segmentation, we adopt the multi-scale cross-entropy loss to restrain the distance between the predicted and target class distributions of pixels at all scales.

\textbf{Multi-scale Reconstruction Loss} We use the $\mathcal{L}_1$ loss to encourage per-pixel reconstruction accuracy, and the perceptual loss~\cite{johnson2016perceptual} to encourage higher-level feature similarity.
\begin{equation}
 \centering
\mathcal{L}_{re}^l(X,\hat{Y}^l) = \parallel X-up(\hat{Y}^l)\parallel _1 +\lambda_p\sum_{n=1}^N\parallel\Psi_n(X)-\Psi_n(up(\hat{Y}^l))\parallel_1,
\end{equation}
where $\Psi_n$ is the activation map of the $n$-th layer, $up(\cdot)$ is the operation to upsample $\hat{Y}^l$ to the same size as $X$, $\lambda_p$ is a trade-off coefficient. We use layered features $relu2\_2$, $relu3\_3$, and $relu4\_3$ in VGG-16 pre-trained on ImageNet to calculate those loss functions.

\textbf{Adversarial Loss} We use a multi-scale PatchGAN \cite{xiao2019cisi} to classify the global and local patches of an image at multiple scales. The multi-scale patch adversarial loss is defined as:
\begin{equation}
\begin{split}
\mathcal{L}_{ad}(X, \hat{Y}) = \sum_{k=1,2,3} (E_{p^k_X\sim X^k}[\log D(p^{k}_X)]+
E_{p^k_{\hat{Y}} \sim \hat{Y}^k}[1-\log D(p^k_{\hat{Y}})]),
\end{split}
 \label{eq:adversarial_loss}
\end{equation}
where $D(\cdot)$ is the discriminator, $p^k_{\hat{Y}}$ and $p^k_{\hat{Y}}$ are the patches in the $k$-th scaled versions of $X$ and $\hat{Y}$.

\textbf{Multi-scale Cross-Entropy Loss} This loss is used to penalize the deviation of $\hat{S}^l$ at each position at every scale.

\begin{equation}
 \centering
\mathcal{L}_{se}^l(S, \hat{S}^l)=-\sum_{i\in S}S_i\log(up(\hat{S}^l)).
 \label{eq:cross_entropy}
\end{equation}
where $i$ indicates each pixel in segmentation map $S$.

\textbf{Final Training Loss} The overall training loss of our network is defined as the weighted sum of the multi-scale reconstruction loss, adversarial loss, and multi-scale cross-entropy loss. 

\begin{equation}
 \centering
\mathcal{L}_{Final}=\sum_{l=0}^4 \mathcal{L}_{re}^l(X,\hat{Y}^l)+\lambda_\alpha\mathcal{L}_{ad}(X, \hat{Y})+\sum_{l=0}^4 \lambda_s\mathcal{L}_{se}^l(S, \hat{S}^l),
 \label{eq:training_loss}
\end{equation}
where $\lambda_\alpha$ and $\lambda_s$ are the weights for the adversarial loss and the multi-scale cross-entropy loss, respectively.

\section{Experiments}

\subsection{Setting}
We evaluate our method on \textbf{Outdoor Scenes} \cite{wang2018recovering} and \textbf{Cityscapes} \cite{cordts2016cityscapes} both with segmentation annotations. \textbf{Outdoor Scenes} contains 9,900 training images and 300 test images belonging to 8 categories. \textbf{Cityscapes} contains 5,000 street view images belonging to 20 categories. In order to enlarge the number of training images of this dataset, we use 2,975 images from the training set and 1,525 images from the test set for training, and test on the 500 images from the validation set. We resize each training image to ensure its minimal height/width to be 256 for \textbf{Outdoor Scenes} and 512 for \textbf{Cityscapes}, and then randomly crop sub-images of size \(256\times256\) as inputs to our model.

We compare our method with the following three representative baselines:

$\bullet$ GC~\cite{yu2019free}: gated convolution for free-form image inpainting, without any auxiliary structural information.

$\bullet$ EC~\cite{nazeri2019edgeconnect}: two-stage inpainting framework with edges as low-level structural information.

$\bullet$ SPG~\cite{song2018spg}: two-stage inpainting framework with a semantic segmentation map as high-level structural information.

In our experiments, we fine-tune the GC and EC models, pre-trained on Places2, on our datasets. We also re-implement and train the model of SPG by ourselves since there is no released code or model. We conduct experiments on both settings of centering and irregular holes. The centering holes are (\(128\times128\) for \textbf{Outdoor Scenes} and \(96\times96\) for \textbf{Cityscapes}), and the irregular masks are obtained from \cite{liu2018image}.

\subsection{Image Inpainting Results}

In this section, we present the results of our model trained on human-annotated segmentation labels. We also verify our model trained on the segmentation labels predicted by a state-of-the-art segmentation model. The results can be found in Section 4.3.

\begin{figure}[t]
	\centering
\footnotesize{
		\begin{tabular}{ccccccc}
			\includegraphics[width=0.14\textwidth]{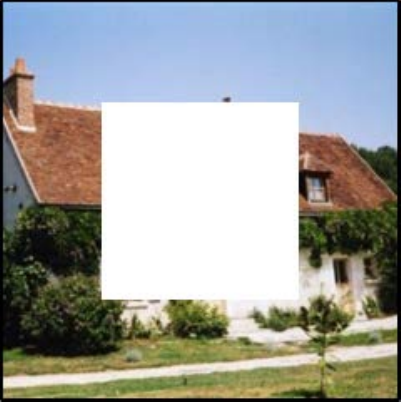} &
			\includegraphics[width=0.14\textwidth]{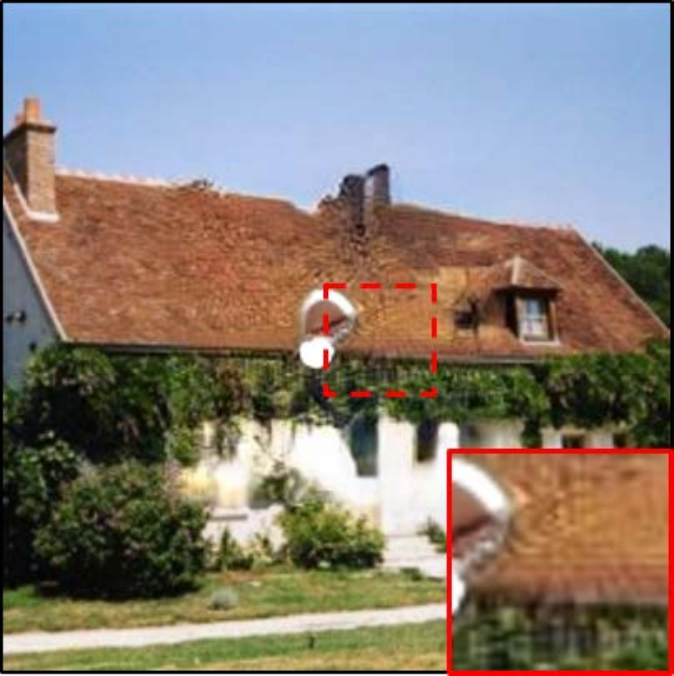} &
			\includegraphics[width=0.14\textwidth]{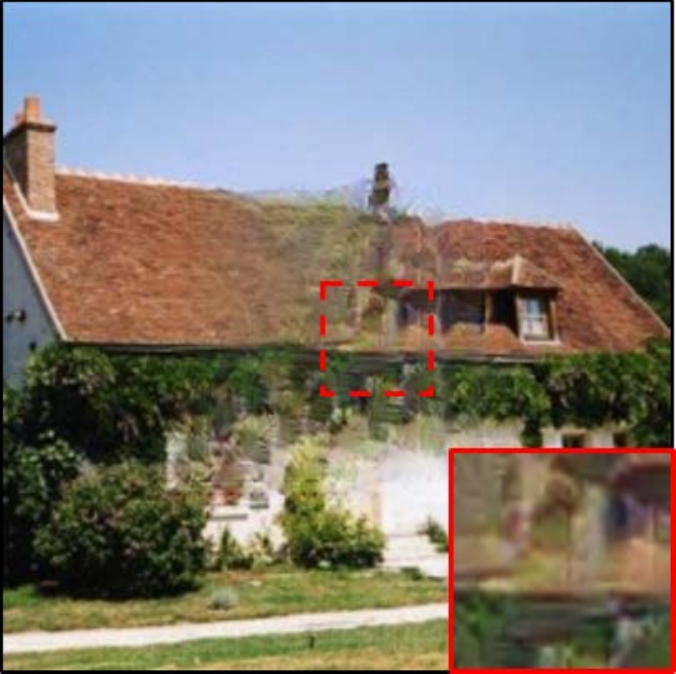} &
			\includegraphics[width=0.14\textwidth]{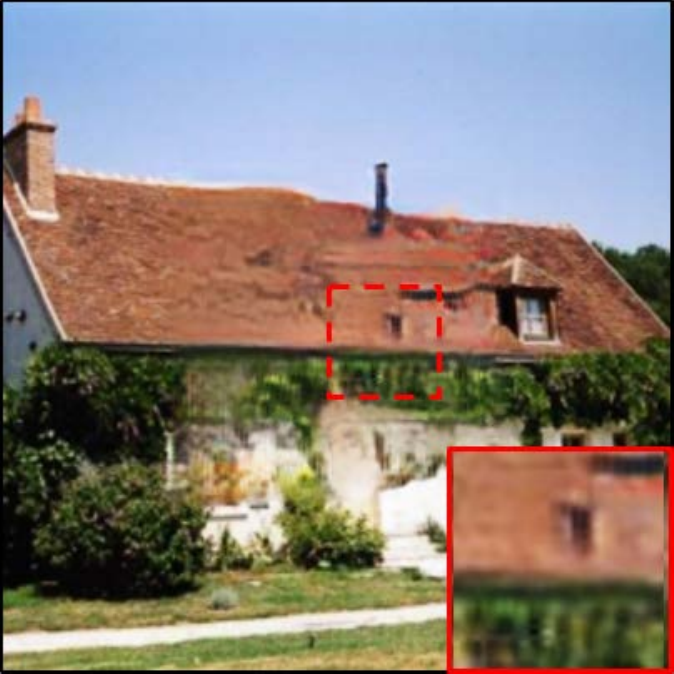} &
			\includegraphics[width=0.14\textwidth]{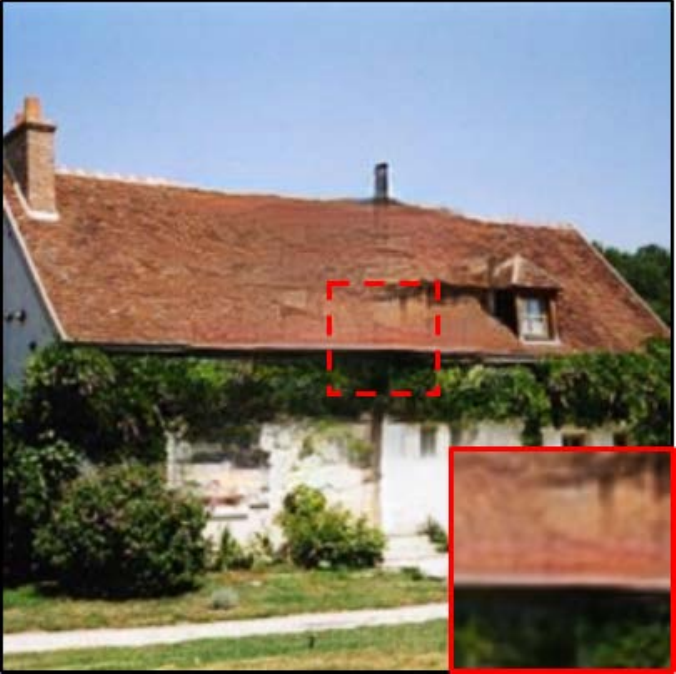} &
			\includegraphics[width=0.14\textwidth]{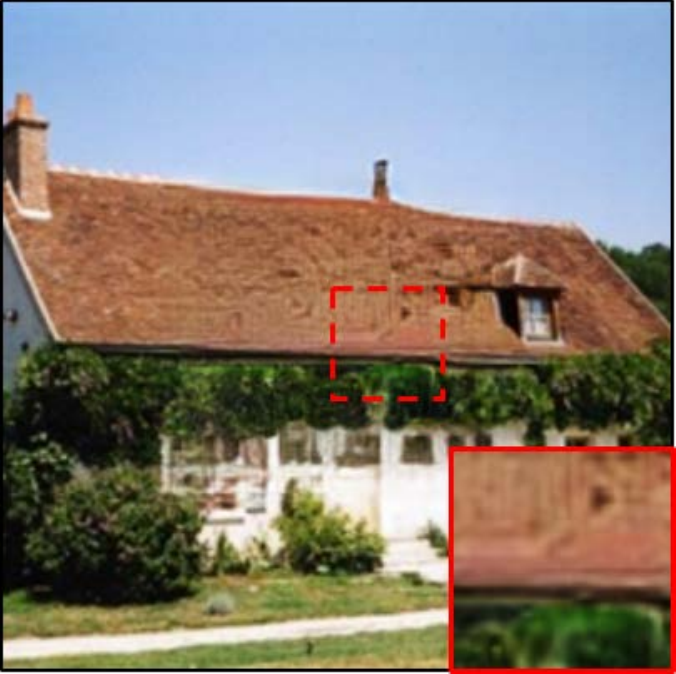} &
			\includegraphics[width=0.14\textwidth]{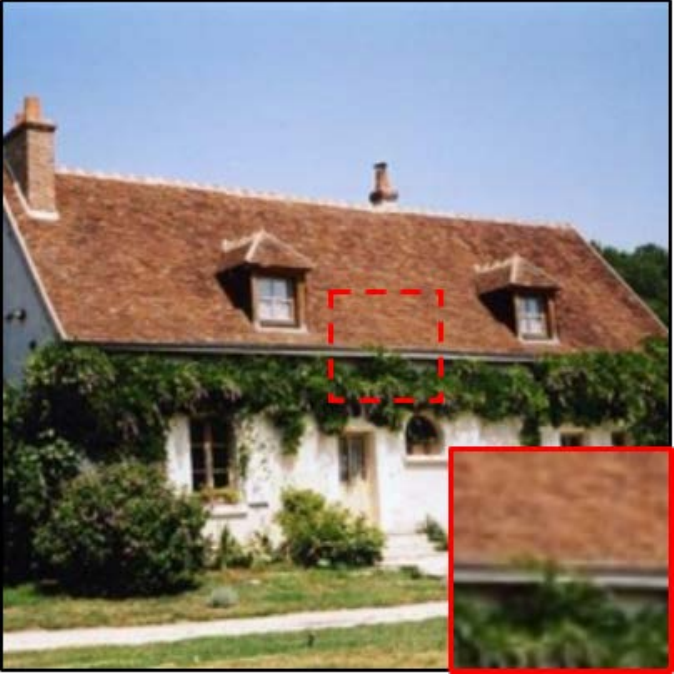} \\
			\includegraphics[width=0.14\textwidth]{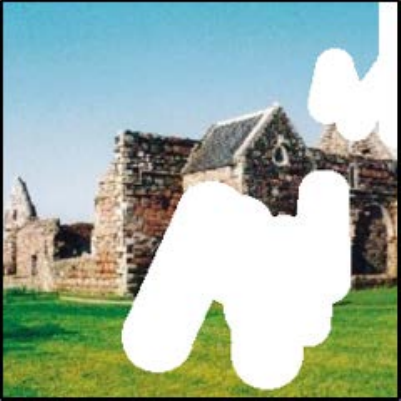} &
			\includegraphics[width=0.14\textwidth]{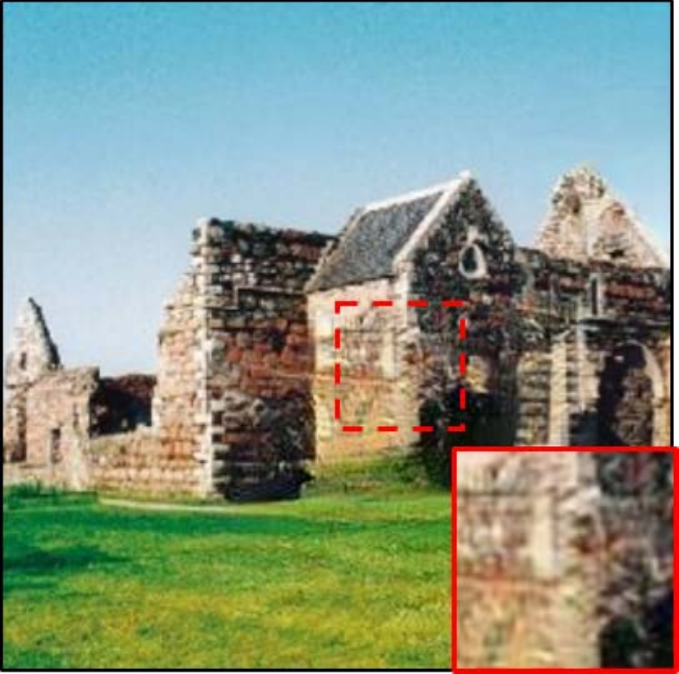} &
			\includegraphics[width=0.14\textwidth]{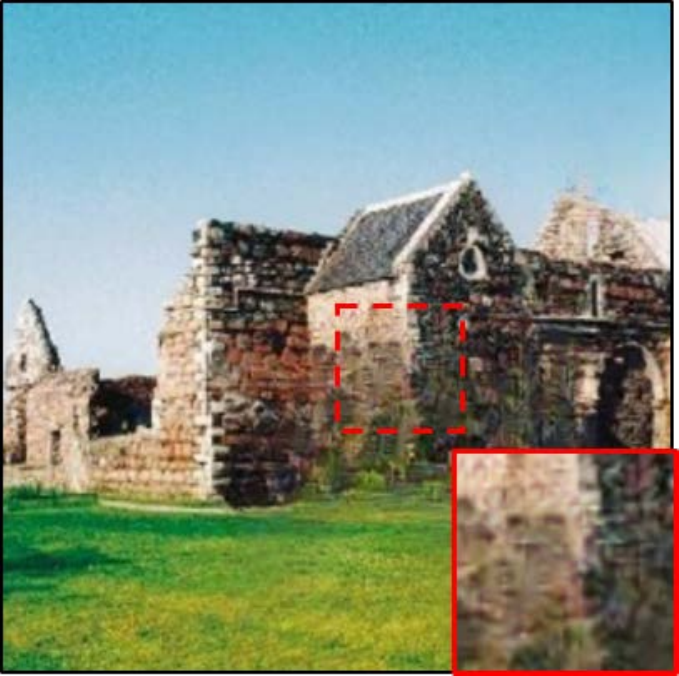} &
			\includegraphics[width=0.14\textwidth]{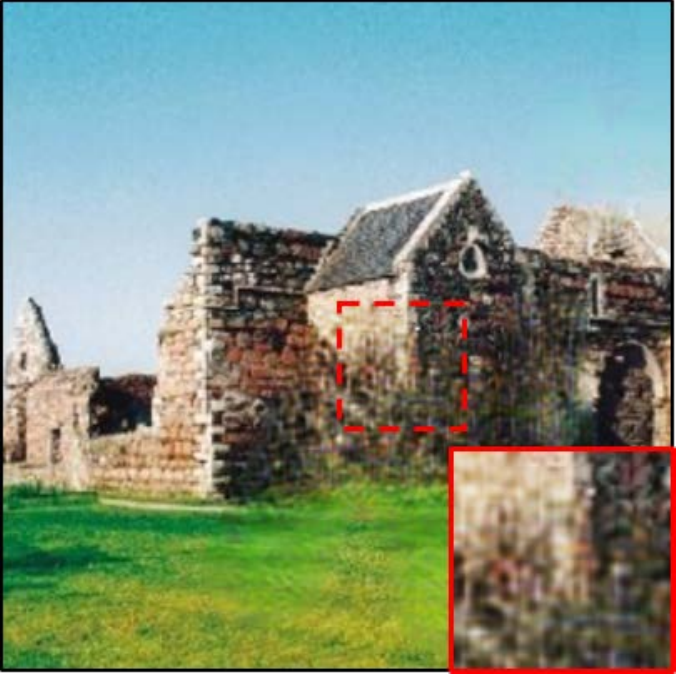} &
			\includegraphics[width=0.14\textwidth]{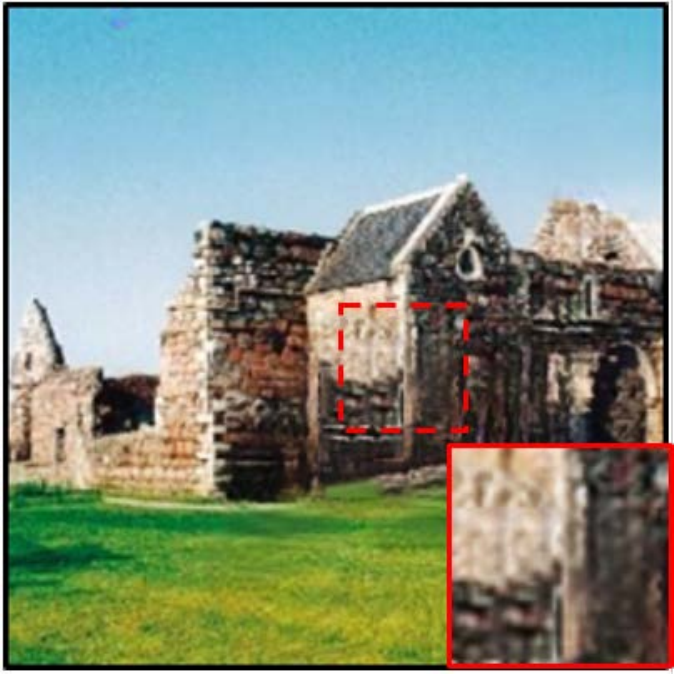} &
			\includegraphics[width=0.14\textwidth]{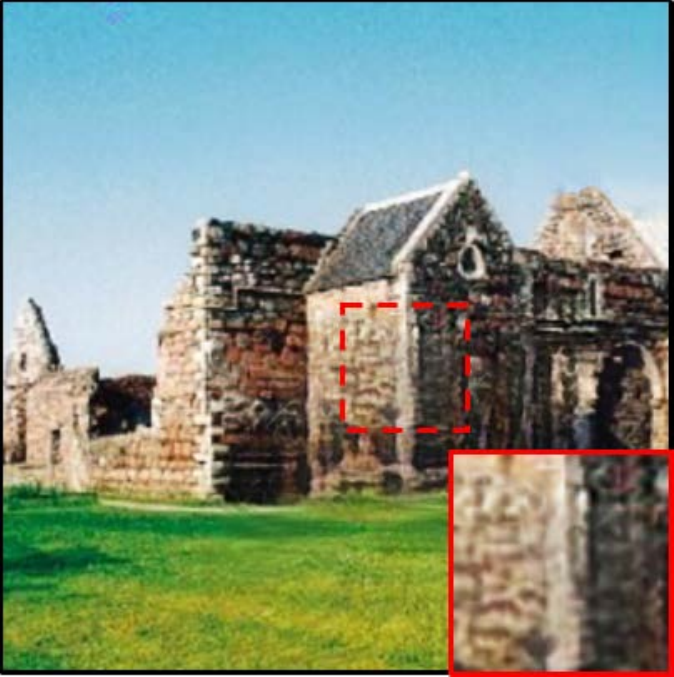} &
			\includegraphics[width=0.14\textwidth]{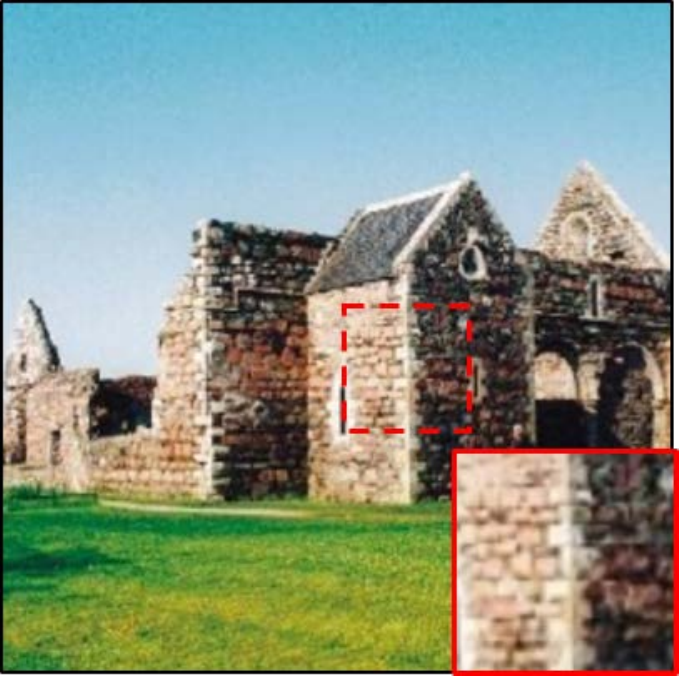} \\
			\includegraphics[width=0.14\textwidth]{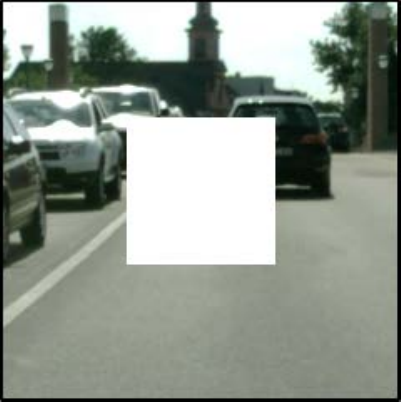} &
			\includegraphics[width=0.14\textwidth]{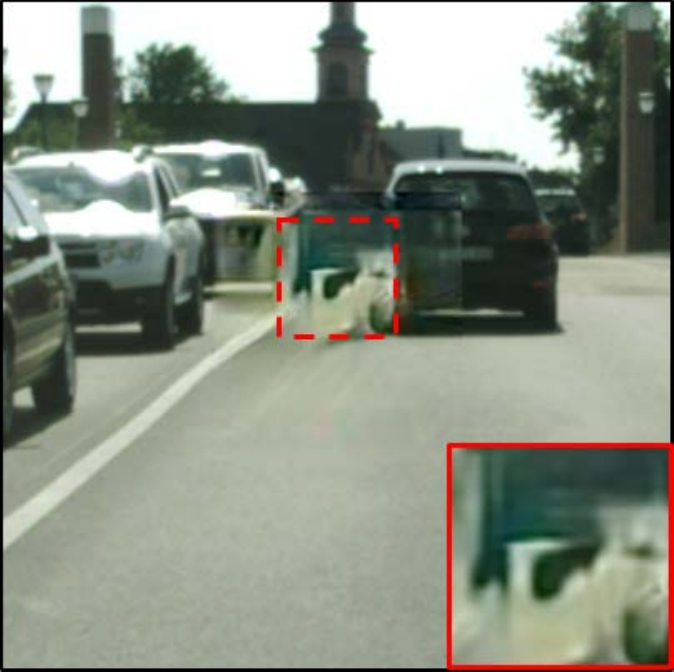} &
			\includegraphics[width=0.14\textwidth]{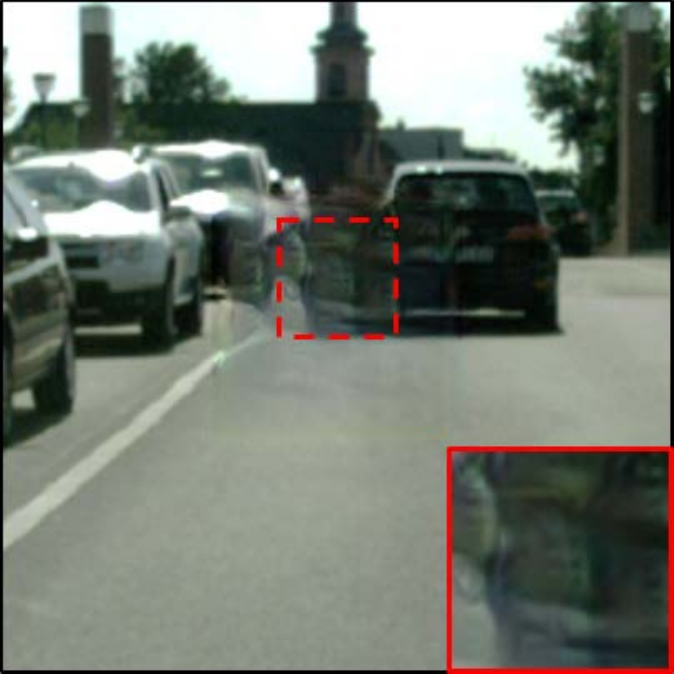} &
			\includegraphics[width=0.14\textwidth]{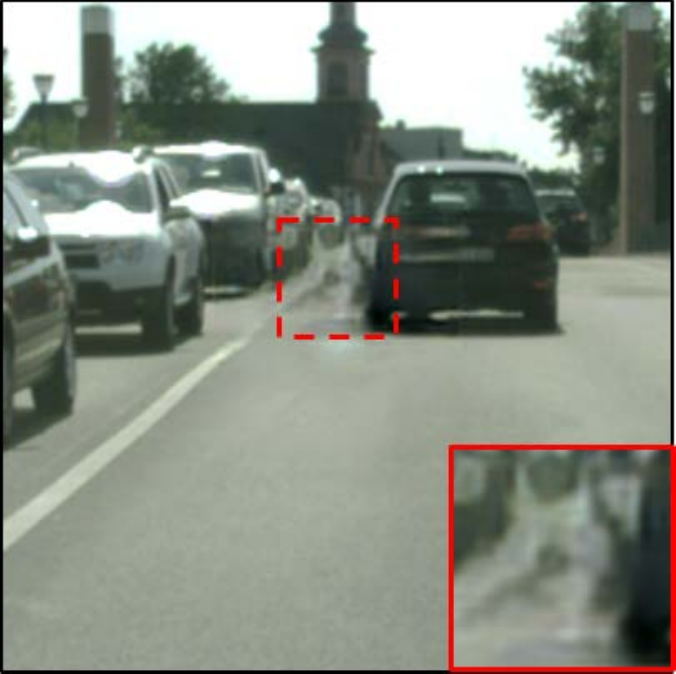} &
			\includegraphics[width=0.14\textwidth]{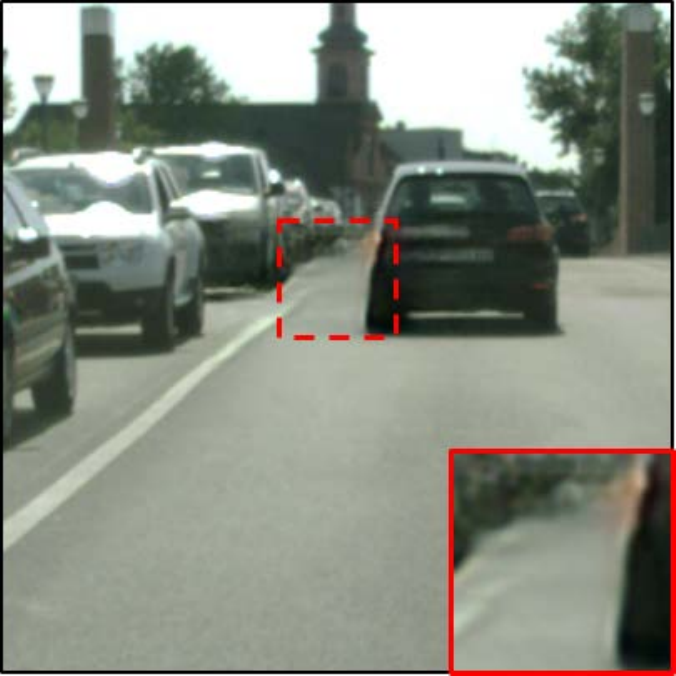} &
			\includegraphics[width=0.14\textwidth]{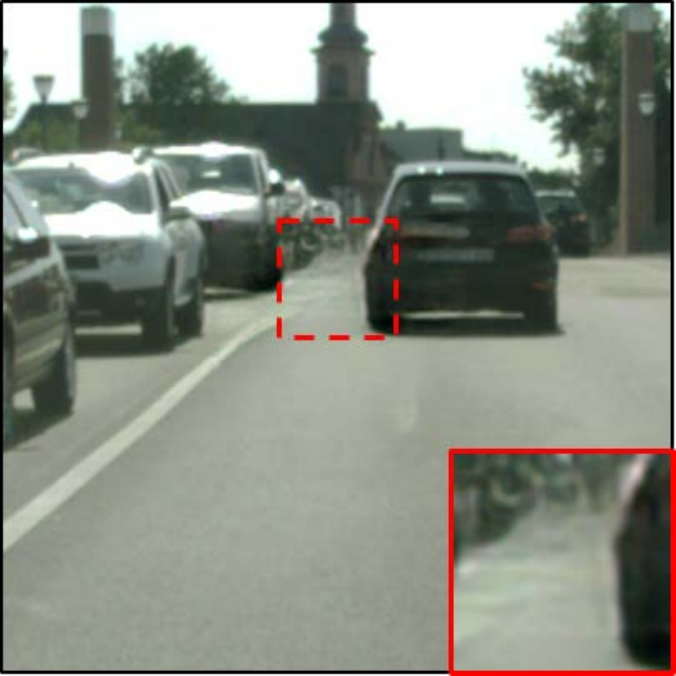} &
			\includegraphics[width=0.14\textwidth]{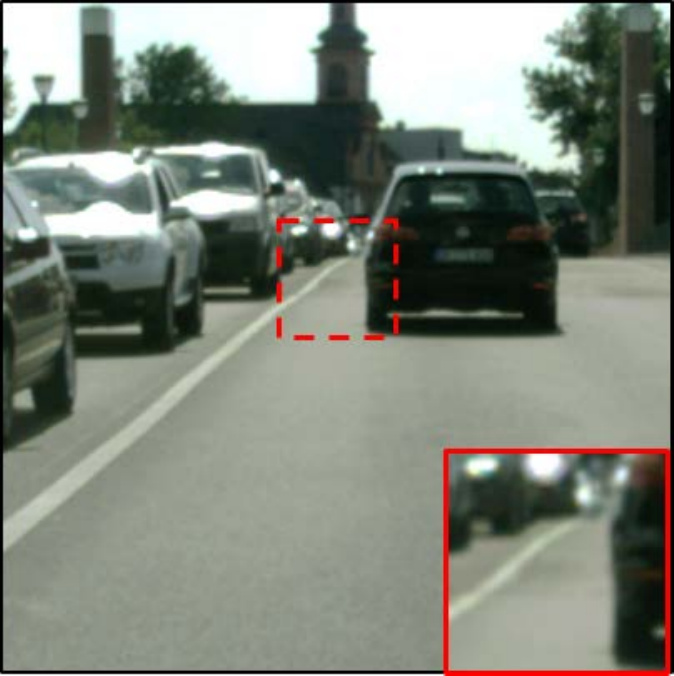} \\
			\includegraphics[width=0.14\textwidth]{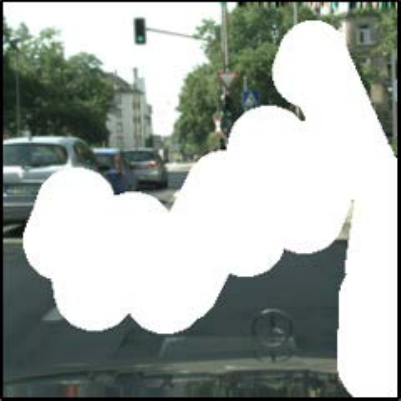} &
			\includegraphics[width=0.14\textwidth]{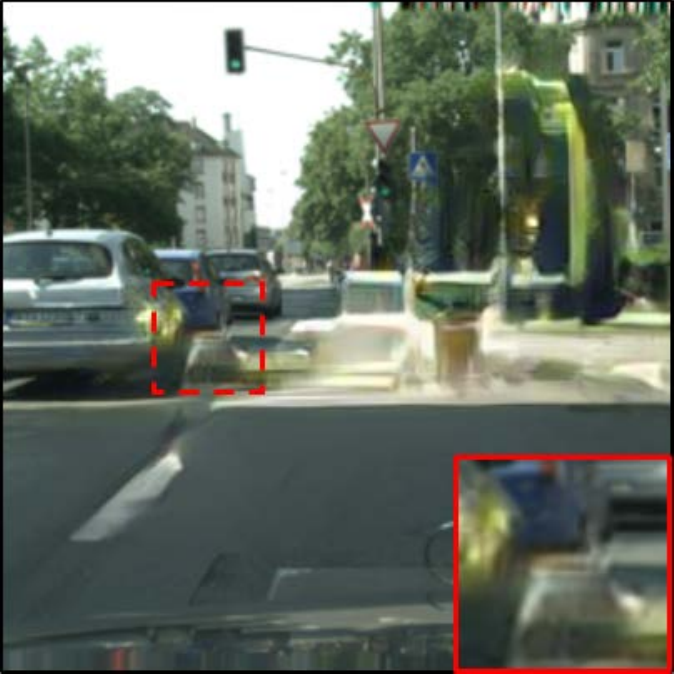} &
			\includegraphics[width=0.14\textwidth]{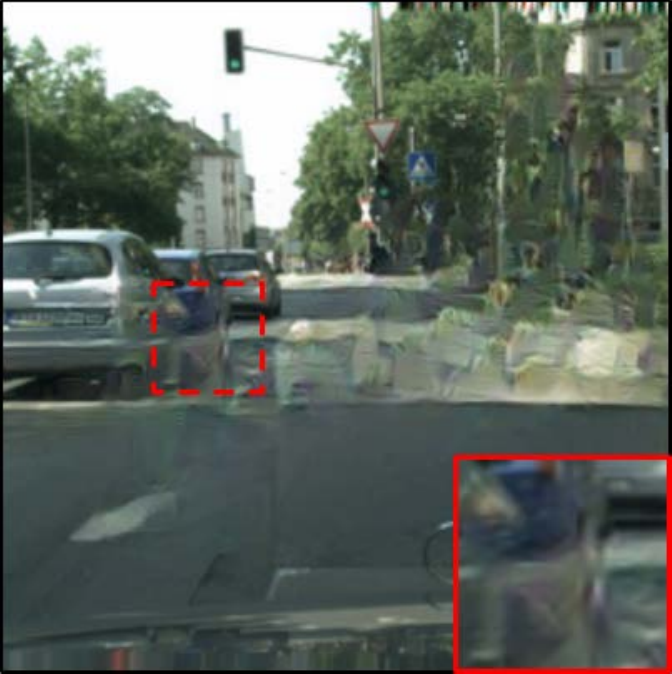} &
			\includegraphics[width=0.14\textwidth]{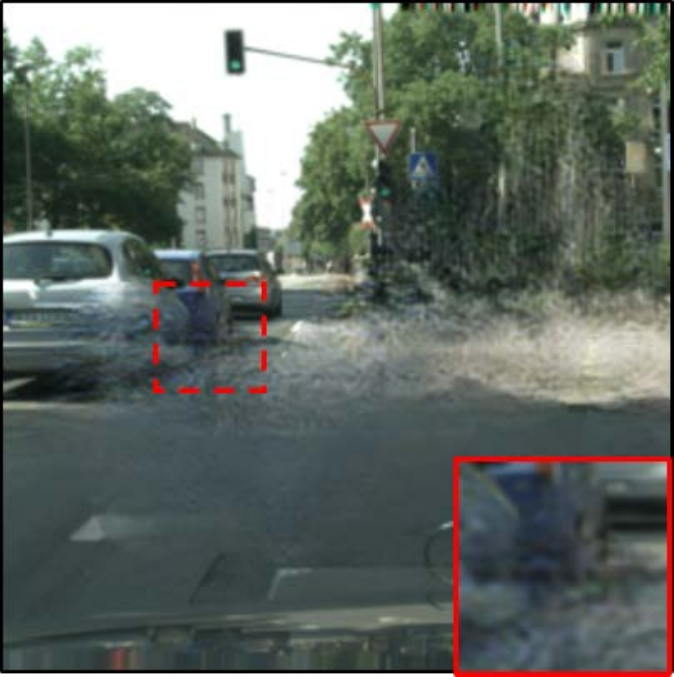} &
			\includegraphics[width=0.14\textwidth]{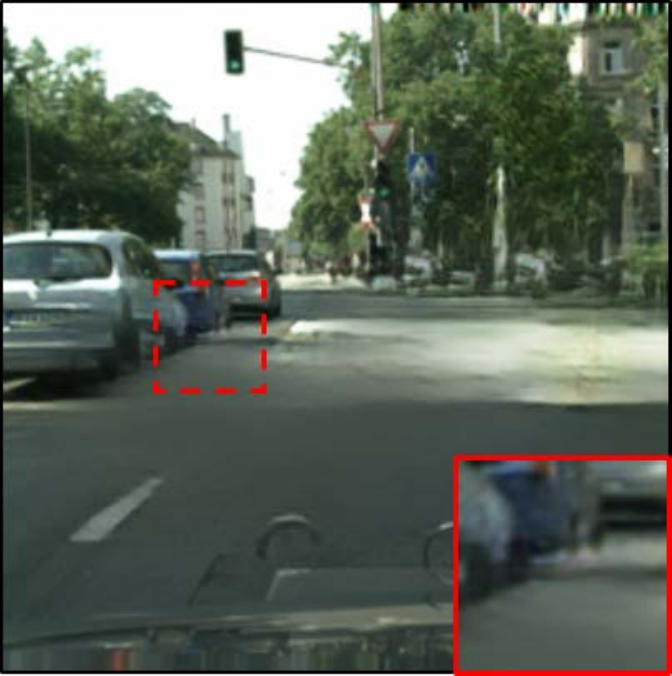} &
			\includegraphics[width=0.14\textwidth]{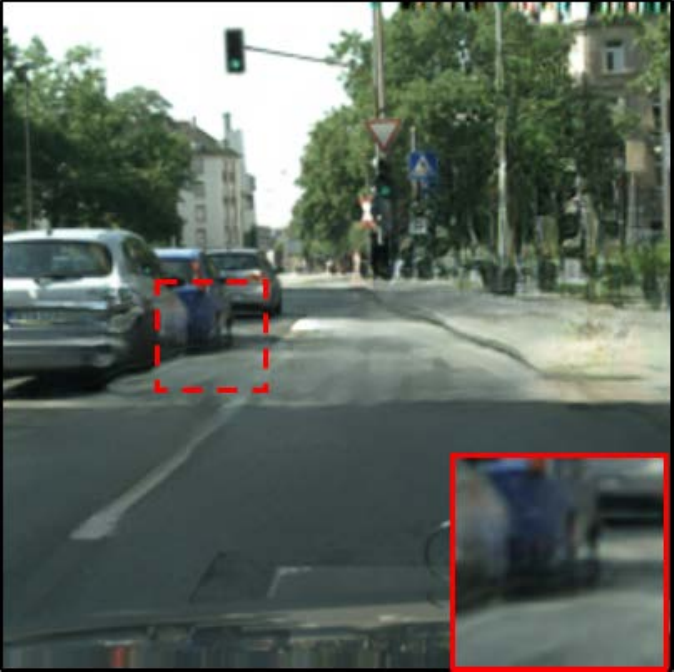} &
			\includegraphics[width=0.14\textwidth]{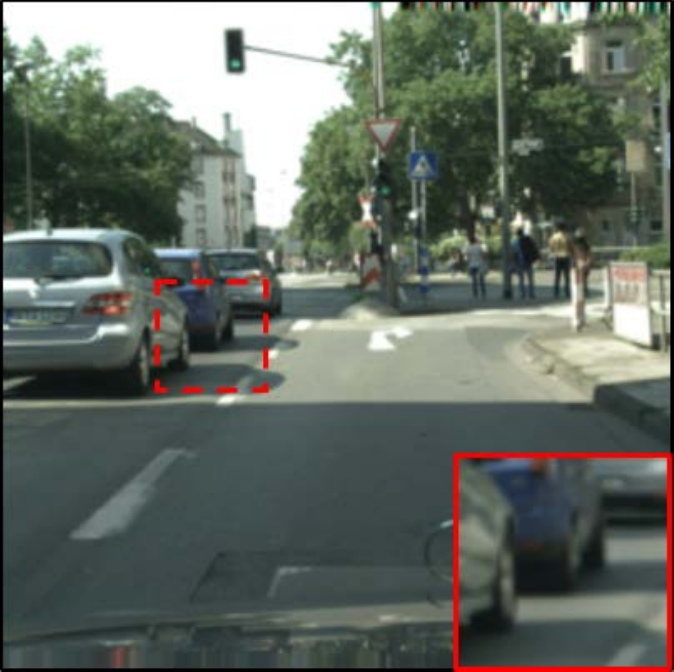} \\
			\scriptsize{Input} & \scriptsize{GC} & \scriptsize{EC} & \scriptsize{SPG} & \scriptsize{SG-Net} & \scriptsize{SGE-Net} & \scriptsize{GT} \\
	\end{tabular}}
   \caption{\scriptsize{Subjective quality comparison of inpainting results on image samples from \textbf{Outdoor Scenes} and \textbf{Cityscapes}. \textbf{GT stands for Ground-Truth}.}}
\label{fig:Overall-results}
\end{figure}

\textbf{Qualitative Comparisons} The subjective visual comparisons of the proposed SG-Net and SGE-Net with the three baselines (GC, EC, SPG) on \textbf{Outdoor Scenes} and \textbf{Cityscapes} are presented in Fig.~\ref{fig:Overall-results}. The corrupted area is simulated by sampling a central hole (\(128\times128\) for \textbf{Outdoor Scenes} and \(96\times96\) for \textbf{Cityscapes}), or 
placing masks with random shapes. As shown in the figure, the baselines usually generate unrealistic shape and textures. The proposed SG-Net generates more realistic textures than the baselines, but still has some flaws at the boundaries since its final result highly depends on the initial inpainting result. The proposed SGE-Net generates better boundaries between semantic regions and more consistent textures than SG-Net and all the baselines, thanks to its evaluation mechanism that can correct the wrongly predicted labels.   

\textbf{Quantitative Comparisons} Table~\ref{tab:exp_stoa} shows the numerical results based on three quality metrics: Peak Signal-to-Noise Ratio (PSNR), Structural Similarity Index (SSIM) and Fréchet Inception Distance (FID) \cite{heusel2017gans}. In general, the proposed SGE-Net  achieves significantly better objective scores than  the baselines, especially in PSNR and SSIM .

\tabcolsep=0.5pt

\begin{table*}[tb]
    \centering
    \caption{\scriptsize{Objective quality comparison of five methods in terms of PSNR, SSIM, and FID on \textbf{Outdoor Scenes} and \textbf{Cityscapes} ($\uparrow$: Higher is better; $\downarrow$: Lower is better). The two best scores are colored in \rc{red} and \bc{blue}, respectively.}}
    \setlength\tabcolsep{5pt}
    \resizebox{\linewidth}{!}{
    \begin{tabular}{l ccc ccc ccc ccc}
    \toprule
    & \multicolumn{6}{c}{Outdoor Scenes} & \multicolumn{6}{c}{Cityscapes} \\
    & \multicolumn{3}{c}{centering holes} & \multicolumn{3}{c}{irregular holes} & \multicolumn{3}{c}{centering holes} & \multicolumn{3}{c}{irregular holes} \\
    \cmidrule(lr){2-4} \cmidrule(lr){5-7} \cmidrule(lr){8-10} \cmidrule(lr){11-13}
    &PSNR$\uparrow$&SSIM$\uparrow$&FID$\downarrow$
    &PSNR$\uparrow$&SSIM$\uparrow$&FID$\downarrow$
    &PSNR$\uparrow$&SSIM$\uparrow$&FID$\downarrow$
    &PSNR$\uparrow$&SSIM$\uparrow$&FID$\downarrow$ \\ \midrule
    
GC~\cite{yu2019free}& 19.06& 0.73 & 42.34
                           & 19.27 & 0.81 &\rc{40.31}
                           & 21.13 & 0.74 & 20.03
                           & 17.42 &0.72 & 40.57 \\
EC~\cite{nazeri2019edgeconnect}& 19.32 &0.76 &\bc{41.25}
                                        & 19.63 & \rc{0.83} &44.31
                                        & 21.71 &0.76 &19.87
                                        & 17.83 &\bc{0.73} &\rc{38.07} \\
SPG~\cite{song2018spg}  & 18.04 & 0.70 & 45.31  
                           & 17.85 &0.74 &50.03
                           & 20.14 & 0.71 & 23.21
                           & 16.01 &0.64 &44.13 \\
\midrule
SG-Net (ours)              & \bc{19.58} & \bc{0.77} & 41.49 
                           & \bc{19.87} & 0.81 &\bc{41.74}
                           & \bc{23.04} &\bc{0.83} &\bc{18.98}
                           & \bc{17.94} &0.64 &41.24  \\ 
\textbf{SGE-Net (ours)}             & \rc{20.53} &\rc{0.81} &\rc{40.67}
                           & \rc{20.02} &\rc{0.83} &42.47
                           & \rc{23.41} &\rc{0.85} &\rc{18.67}
                           & \rc{18.03} &\rc{0.75} &\bc{39.93}  \\ 
    \bottomrule
    \end{tabular}
    }
    \label{tab:exp_stoa}
\end{table*}

\textbf{User Study}
We conduct the user study on 80 images randomly selected from both datasets. In total, 24 subjects with some background of image processing are involved to rank the subjective visual qualities of images completed by four inpainting methods (GC, EC, SPG, and our SGE-Net). As shown in Table~\ref{tab:user}, the study shows that $67.4\%$ of subjects (1295 out of 1920 comparisons), $70.7\%$ and $73.2\%$ preferred our results over GC, EC, and SPG, respectively. Hence, our method outperforms the other methods.


Since our method mainly focuses on completing mixed scenes with multiple semantics, we also verify its performance on images with different scene complexities. We conduct this analysis by dividing all 80 images into three levels of semantic complexities: 1) low-complexity scenes containing 27 images with 1--2 semantics; 2) moderate-complexity scenes containing 32 images with 3--4 semantics; 3) high-complexity scenes containing 21 images with more than 4 semantics. As shown in Table~\ref{tab:user},  compared to the baselines, while our method achieves generally better results than the baselines for the simple- to moderate-complexity scenes (about from $60\%$ to $70\%$), the preference rate increases significantly for the complex scenes (from $77.6\%$ to $83.7\%$). This verifies that our method is particularly powerful for completing mixed-scene images with multiple semantics, thanks to its mechanism for understanding and updating the semantics. 

\begin{figure}[tb]
    \begin{minipage}[c]{.49\linewidth}
    \centering
	\footnotesize{
		\begin{tabular}{ccccc}
			\includegraphics[width=0.2\columnwidth]{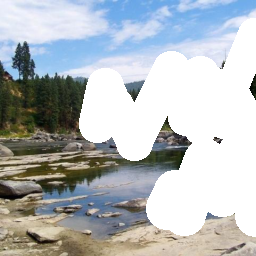} &
			\includegraphics[width=0.2\columnwidth]{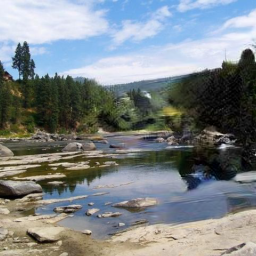} &
			\includegraphics[width=0.2\columnwidth]{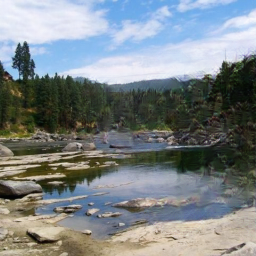} &
			\includegraphics[width=0.2\columnwidth]{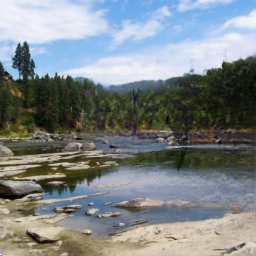} &
			\includegraphics[width=0.2\columnwidth]{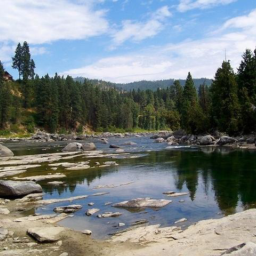} \\
			\includegraphics[width=0.2\columnwidth]{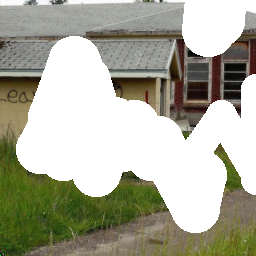} &
			\includegraphics[width=0.2\columnwidth]{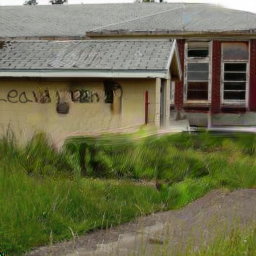} &
			\includegraphics[width=0.2\columnwidth]{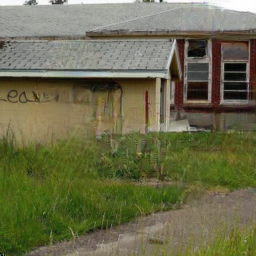} &
			\includegraphics[width=0.2\columnwidth]{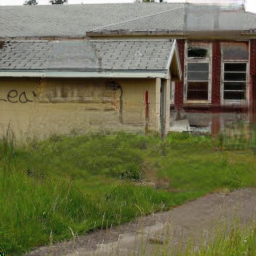} &
			\includegraphics[width=0.2\columnwidth]{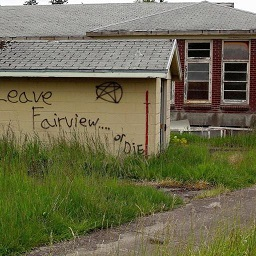} \\
			
			\scriptsize{Input} & \scriptsize{GC} & \scriptsize{EC} & \scriptsize{SGE-Net} & \scriptsize{GT}\\
	\end{tabular}}
	\linespread{1}
	\caption{\scriptsize{Subjective quality comparison on image samples from \textbf{Places2}.}}
    \label{fig:places2}
        \end{minipage}
    \hspace{1mm}
    \begin{minipage}[c]{.49\linewidth}
    	\centering
	\footnotesize{
		\begin{tabular}{ccccc}
			\includegraphics[width=0.2\columnwidth]{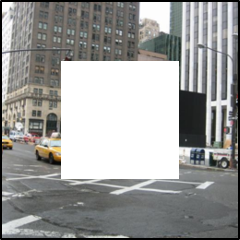} &
			\includegraphics[width=0.2\columnwidth]{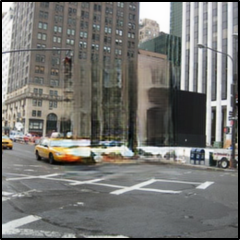} &
			\includegraphics[width=0.2\columnwidth]{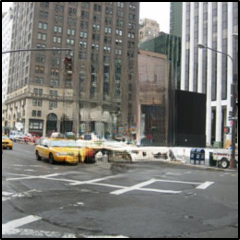} &
			\includegraphics[width=0.2\columnwidth]{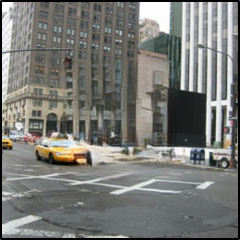} &
			\includegraphics[width=0.2\columnwidth]{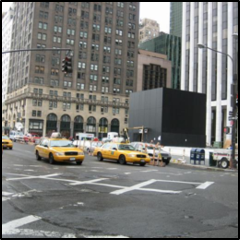} \\
			\includegraphics[width=0.2\columnwidth]{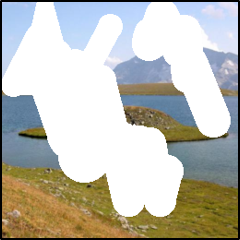} &
			\includegraphics[width=0.2\columnwidth]{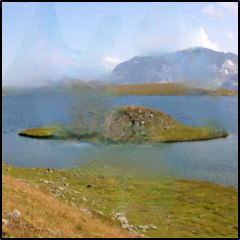} &
			\includegraphics[width=0.2\columnwidth]{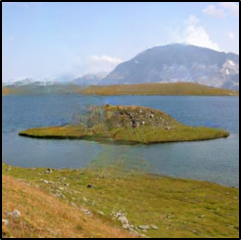} &
			\includegraphics[width=0.2\columnwidth]{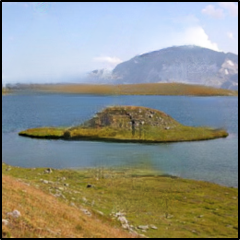} &
			\includegraphics[width=0.2\columnwidth]{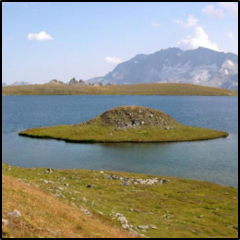} \\
			\scriptsize{Input} & \scriptsize{Basic-Net} & \scriptsize{SG-Net} & \scriptsize{SGE-Net} & \scriptsize{GT}  \\
	\end{tabular}}    
	\captionof{figure}{\scriptsize{Subjective visual quality comparisons on the effects of SGIM and SCEM.}}
    \label{fig:SGIMandSEM}
    \end{minipage}
\end{figure}

\textbf{Additional Results on Places2}
For fair comparison, we also test our method on \textbf{Places2} \cite{zhou2017places} to verify that SGE-Net can be applied to images without segmentation annotations. \textbf{Places2} was used for evaluation by both GC and EC. It contains images with similar semantic scenes to \textbf{Outdoor Scenes}. Therefore, we use our model trained on \textbf{Outdoor Scenes} to complete the images with similar scenes in \textbf{Places2}. The subjective results in Fig.~\ref{fig:places2} show that SGE-Net is still able to generate proper semantic structures, owing to the introduction of the semantic segmentation, which provides the prior knowledge about the scenes. 

\tabcolsep=0.5pt
\begin{table}[tb]
	\centering
	    \caption{\scriptsize{Preference percentage matrix (\%) of different scene complexities on \textbf{Outdoor Scenes} and  \textbf{Cityscapes} datasets. Overall, low complexity, moderate complexity, and high complexity are colored in black, \gc{green}, \bc{blue} and \rc{red}, respectively. }}
     \setlength\tabcolsep{4pt}
    \resizebox{\linewidth}{!}{
    \begin{tabular}{lcccc}
    \toprule
                & \scriptsize{GC~\cite{yu2019free}}   & \scriptsize{EC~\cite{nazeri2019edgeconnect}} & \scriptsize{SPG~\cite{song2018spg}}      & \scriptsize{\textbf{SGE-Net (ours)}}           \\ \midrule
     \scriptsize{GC~\cite{yu2019free}}  & --  
                &\scriptsize{(46.7)/\gc{41.5}/\bc{47.7}/\rc{52.0} }
                &\scriptsize{(58.1)/\gc{57.3}/\bc{59.8}/\rc{56.7} } 
                &\scriptsize{(32.6)/\gc{37.8}/\bc{34.8}/\rc{22.4}}\\
\scriptsize{EC~\cite{nazeri2019edgeconnect}}     &\scriptsize{(53.3)/\gc{58.5}/\bc{52.3}/\rc{48.0}}
                & --  &\scriptsize{(70.1)/\gc{68.7}/\bc{69.3}/\rc{73.0} }
                &\scriptsize{(29.3)/\gc{35.0}/\bc{31.8}/\rc{18.1}}  \\
    \scriptsize{SPG~\cite{song2018spg}}     &\scriptsize{(41.9)/\gc{42.7}/\bc{40.2}/\rc{43.3}} 
                &\scriptsize{(29.9)/\gc{31.3}/\bc{30.7}/\rc{27.0}} 
                & -- 
                &\scriptsize{(26.8)/\gc{32.7}/\bc{28.8}/\rc{16.3}} \\
    \scriptsize{\textbf{SGE-Net (ours)}}         &\scriptsize{(67.4)/\gc{62.2}/\bc{65.2}/\rc{77.6} }
                &\scriptsize{(70.7)/\gc{65.0}/\bc{68.2}/\rc{81.9}}
                &\scriptsize{(73.2)/\gc{67.3}/\bc{71.2}/\rc{83.7} } 
                & --  \\
    \bottomrule
    \end{tabular}
    }
    \label{tab:user}
\end{table}

\tabcolsep=0.5pt

\subsection{Ablation Study}
\textbf{Effectiveness of SGIM and SCEM}
In the proposed networks, the two core components of our method, semantic-guided inference and segmentation confidence evaluation, are implemented by SGIM and SCEM, respectively. In order to investigate their effectiveness, we conduct an ablation study on three variants: a) Basic-Net (without SGIM and SCEM); b) SG-Net (with SGIM but without SCEM); and c) SGE-Net (with both SGIM and SCEM).

The visual and numeric comparisons on \textbf{Outdoor Scenes} are shown in Fig.~\ref{fig:SGIMandSEM} and Table~\ref{tab:SGIMandSEM}. In general, the inpainting performance increases with the added modules. Specifically, the multi-scale semantic-guided interplay framework does a good job for generating detailed contents, and the semantic segmentation map helps learn a more accurate layout of a scene. With SGIM, the spatial adaptive normalization helps generate more realistic textures based on the semantic priors. Moreover, SCEM makes further improvements on completing structures and textures (fourth column in Fig.~\ref{fig:SGIMandSEM}) by coarse-to-fine optimizing the semantic contents across scales. 

\begin{wraptable}{r}{0.4\textwidth}
\centering
    \caption{\scriptsize{Objective quality comparison on the performances of SGIM and SCEM in terms of three metrics.}}
    \setlength\tabcolsep{2pt}
    \resizebox{\linewidth}{!}{
    \begin{tabular}{lccccc}
    \toprule
              & SGIM & SCEM & PSNR$\uparrow$&SSIM$\uparrow$&FID$\downarrow$ \\ \midrule
    Basic-Net & \xmark     & \xmark       &19.14 & 0.71 & 43.43 \\
    SG-Net    & \cmark     & \xmark       & 19.58 & 0.77 &41.49 \\
    \textbf{SGE-Net}   & \cmark     & \cmark       & \bf{20.53} & \bf{0.81} & \bf{40.67} \\
    \bottomrule
    \end{tabular}
    }
    \label{tab:SGIMandSEM}

\end{wraptable}

\begin{figure}[tb]
	\centering
	\footnotesize{
		\begin{tabular}{cccccccc}
			\includegraphics[width=0.122\textwidth]{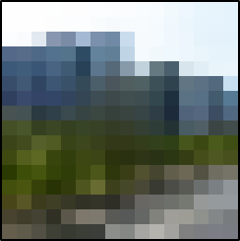} &
			\includegraphics[width=0.122\textwidth]{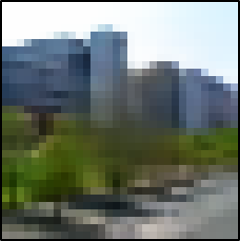} &
			\includegraphics[width=0.122\textwidth]{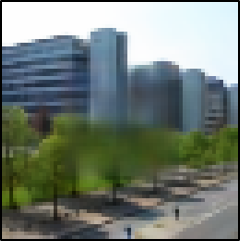} &
			\includegraphics[width=0.122\textwidth]{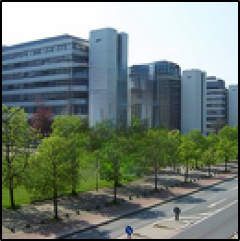} &
			\includegraphics[width=0.122\textwidth]{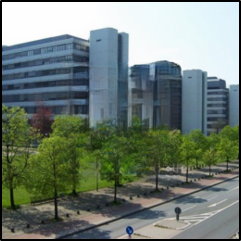} &
			\includegraphics[width=0.122\textwidth]{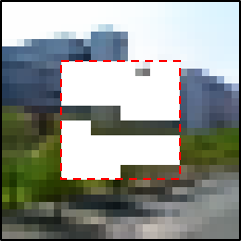} &
			\includegraphics[width=0.122\textwidth]{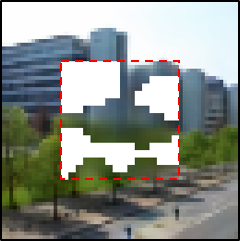} & 
			\includegraphics[width=0.122\textwidth]{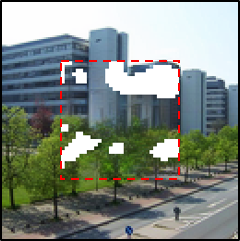} \\
			\includegraphics[width=0.122\textwidth]{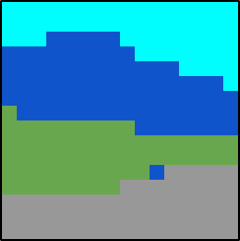} &
			\includegraphics[width=0.122\textwidth]{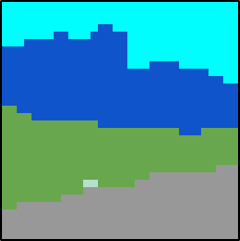} &
			\includegraphics[width=0.122\textwidth]{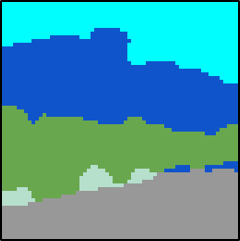} &
			\includegraphics[width=0.122\textwidth]{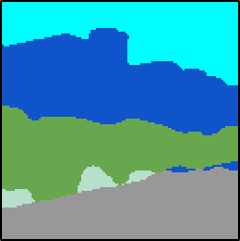} &
			\includegraphics[width=0.122\textwidth]{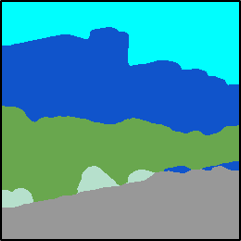} &
			\includegraphics[width=0.122\textwidth]{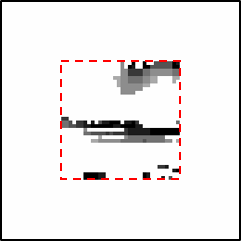} &
			\includegraphics[width=0.122\textwidth]{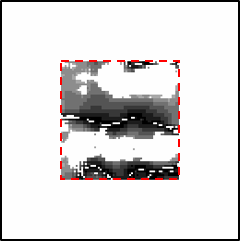} &
			\includegraphics[width=0.122\textwidth]{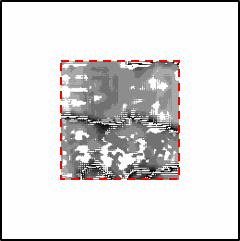} \\
			\scriptsize{scale 4} & \scriptsize{scale 3} & \scriptsize{scale 2} & \scriptsize{scale 1} & \scriptsize{final} & \scriptsize{$\hat{S}^{4-3}_{p_{max}}$} & \scriptsize{$\hat{S}^{3-2}_{p_{max}}$} & \scriptsize{$\hat{S}^{2-1}_{p_{max}}$}\\
	\end{tabular}}
   \caption{\scriptsize{Illustration of multi-scale progressive refinement with SGE-Net. From left to right of the first 5 columns: the inpainted images (top row) and the segmentation maps (bottom row) from scale 4 to scale 1 and the final result. The last 3 columns show the the reliability maps (top row) and the confidence score maps (bottom row)  of the inpainted area across scales (e.g., $\hat{S}_{P_{max}}^{4-3}$ shows  the confidence score increases from scale 4 to 3).}}
\label{fig:scale-change}
\end{figure}

To further verify the effectiveness of SCEM, we visualize a corrupted image and its segmentation maps derived from all decoding scales. As shown in the first five columns of Fig.~\ref{fig:scale-change}, the multi-scale progressive-updating mechanism gradually refines the  detailed textures as illustrated in the images and the segmentation maps at different scales. The last three columns of the top row show that the region of the unreliable mask gradually shrinks as well. Correspondingly, the bottom row shows the increase of the confidence scores of segmentation maps from left to right (e.g., $\hat{S}_{P_{max}}^{4-3}$ showing the increase of  the confidence score from scale 4 to scale 3). The proportion of the white region, which roughly indicates unreliable labels, also decreases significantly from left to right. The result evidently demonstrates the benefits of SCEM in strengthening the semantic correctness of contextual features. 

\tabcolsep=0.5pt

\begin{figure}[tb]
	\centering
\footnotesize{
		\begin{tabular}{ccccccccccc}
			\includegraphics[width=0.093\textwidth]{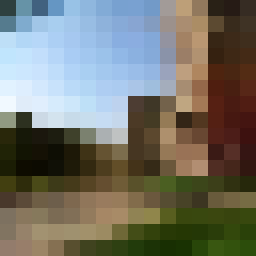} &
			\includegraphics[width=0.093\textwidth]{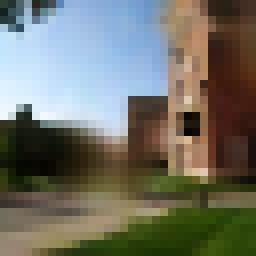} &
			\includegraphics[width=0.093\textwidth]{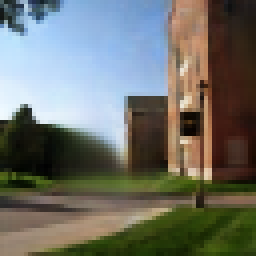} &
			\includegraphics[width=0.093\textwidth]{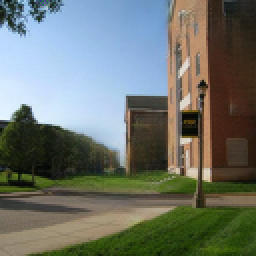} &
			\includegraphics[width=0.093\textwidth]{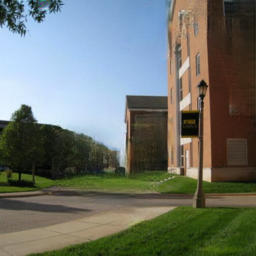}&
			\tiny{~}&
			\includegraphics[width=0.093\textwidth]{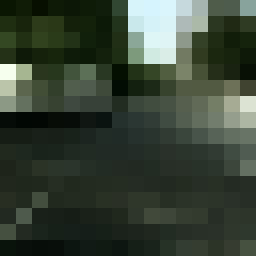} &
			\includegraphics[width=0.093\textwidth]{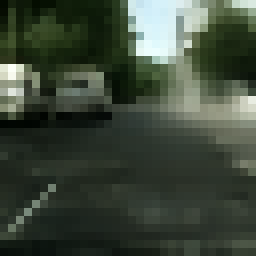} &
			\includegraphics[width=0.093\textwidth]{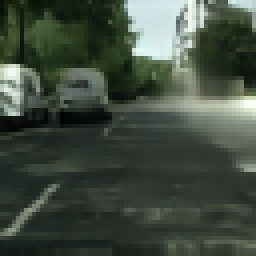} &
			\includegraphics[width=0.093\textwidth]{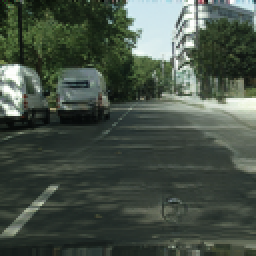} &
			\includegraphics[width=0.093\textwidth]{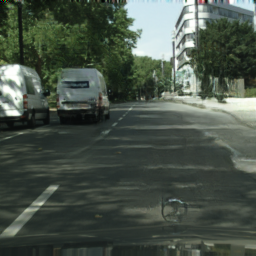} \\
			\includegraphics[width=0.093\textwidth]{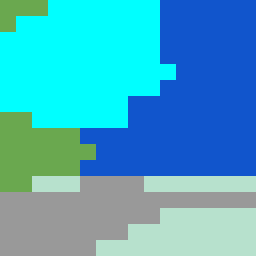} &
			\includegraphics[width=0.093\textwidth]{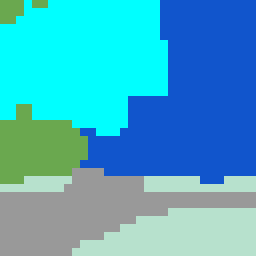} &
			\includegraphics[width=0.093\textwidth]{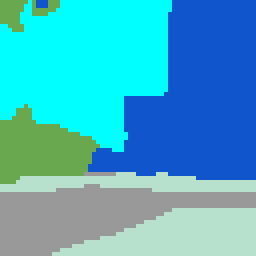} &
			\includegraphics[width=0.093\textwidth]{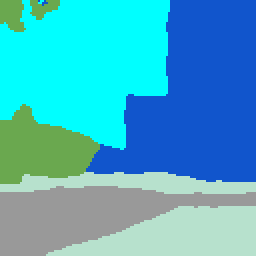} &
			\includegraphics[width=0.093\textwidth]{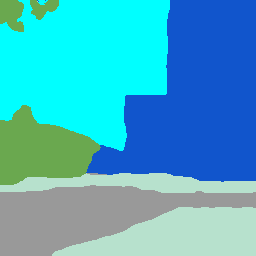}&
			\tiny{~}&
			\includegraphics[width=0.093\textwidth]{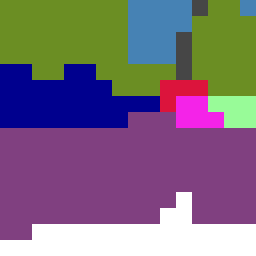} &
			\includegraphics[width=0.093\textwidth]{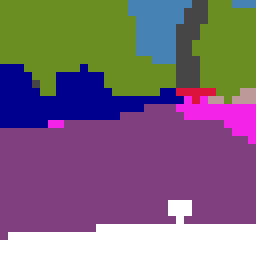} &
			\includegraphics[width=0.093\textwidth]{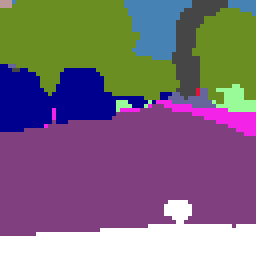} &
			\includegraphics[width=0.093\textwidth]{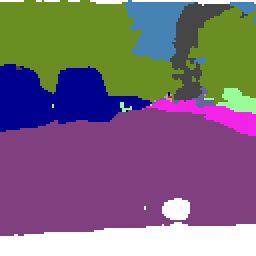} &
			\includegraphics[width=0.093\textwidth]{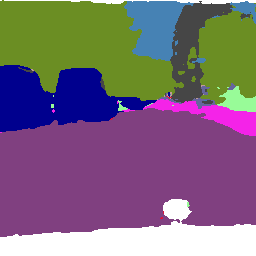}\\
			\includegraphics[width=0.093\textwidth]{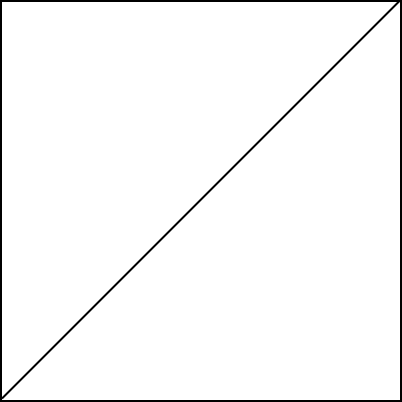} &
			\includegraphics[width=0.093\textwidth]{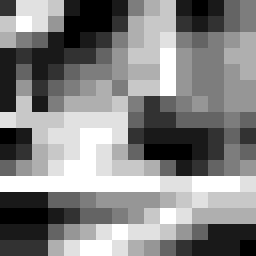} &
			\includegraphics[width=0.093\textwidth]{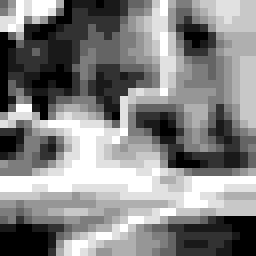} &
			\includegraphics[width=0.093\textwidth]{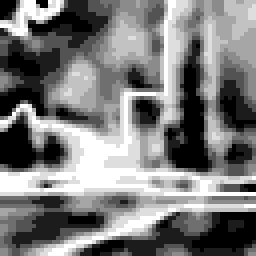} &
			\includegraphics[width=0.093\textwidth]{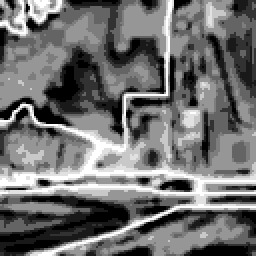} &
			\tiny{~}&
			\includegraphics[width=0.093\textwidth]{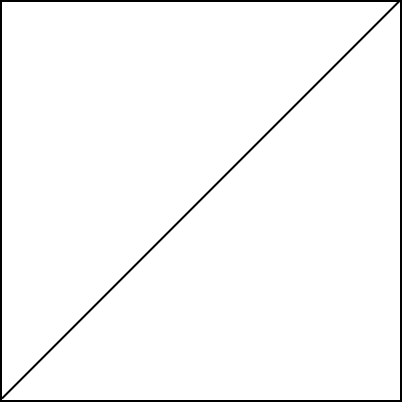} &
			\includegraphics[width=0.093\textwidth]{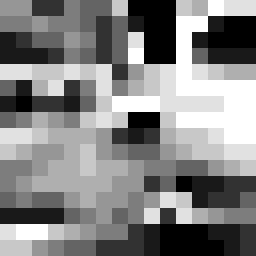} &
			\includegraphics[width=0.093\textwidth]{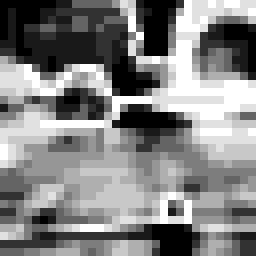} &
			\includegraphics[width=0.093\textwidth]{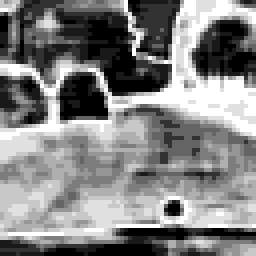} &
			\includegraphics[width=0.093\textwidth]{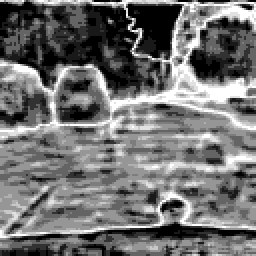} \\
			\includegraphics[width=0.093\textwidth]{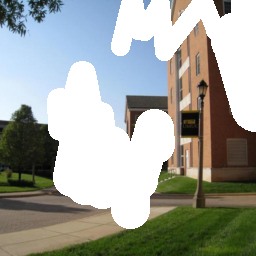} &
			\includegraphics[width=0.093\textwidth]{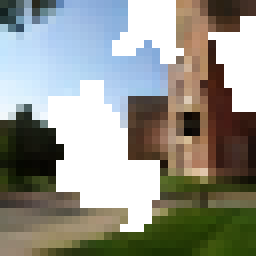} &
			\includegraphics[width=0.093\textwidth]{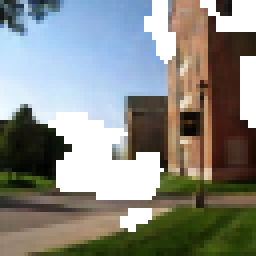} &
			\includegraphics[width=0.093\textwidth]{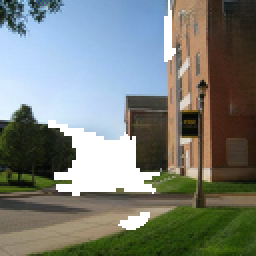} &
			\includegraphics[width=0.093\textwidth]{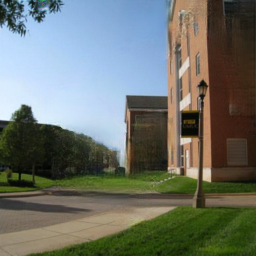} &
			\tiny{~}&
			\includegraphics[width=0.093\textwidth]{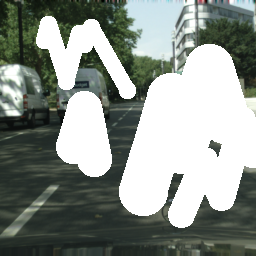} &
			\includegraphics[width=0.093\textwidth]{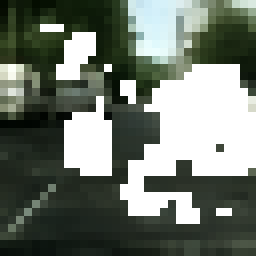} &
			\includegraphics[width=0.093\textwidth]{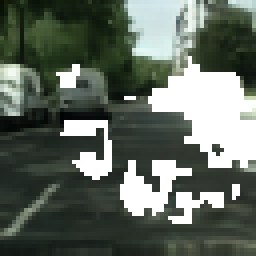} &
			\includegraphics[width=0.093\textwidth]{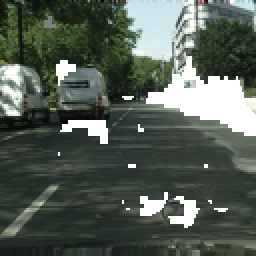} &
			\includegraphics[width=0.093\textwidth]{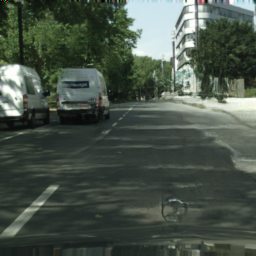} \\
			\scriptsize{scale 4} & \scriptsize{scale 3} & \scriptsize{scale 2} & \scriptsize{scale 1} & \scriptsize{final} & 
			\tiny{~}& \scriptsize{scale 4} & \scriptsize{scale 3} & \scriptsize{scale 2} & \scriptsize{scale 1} & \scriptsize{final} \\
			\multicolumn{5}{c}{\scriptsize{(a) \textbf{Outdoor Scenes}}} & \tiny{~} & \multicolumn{5}{c}{\scriptsize{(b) \textbf{Cityscapes}}}
	\end{tabular}}
		    \linespread{1}
   \caption{\scriptsize{Correspondence between the confidence score value and the reliability of inpainted image content. Row 1: Inpainted image. Row 2: Predicted segmetation map. Row 3: the confidence score map (darker color means higher confidence score, and vice versa). Row 4: unreliable pixel map (white pixels indicate unreliable pixels). Since the map at scale 4 is the same as the input mask, we put the input image for better comparison.}}
 \label{fig:Mask-update}
    \end{figure}

\textbf{Justification of Segmentation Confidence Scoring}
During the progressive refinement of image inpainting and semantic segmentation, the semantic evaluation mechanism of SCEM is based on the assumption that the pixel-wise confidence scores from the segmentation possibility map can well reflect the correctness of the inpainted pixel values. Here we attempt to justify this assumption. Some examples from both datasets are shown in Fig.~\ref{fig:Mask-update}. 
It can be seen that (except for the confidence scores at the region boundaries): a) The low confidence scores (the white area in row 3) usually appear in the mask area, indicating that the scores reasonably well reflects the reliability of inpainted image content; b) the confidence score becomes higher when the scale goes finer, and correspondingly the area of unreliable pixels reduces, meaning that our method can progressively refine the context feature towards correct inpainting.

\tabcolsep=0.5pt

\begin{figure}[htb]
    \begin{minipage}[c]{.48\linewidth}
      \begin{center}
    \includegraphics[width=\columnwidth]{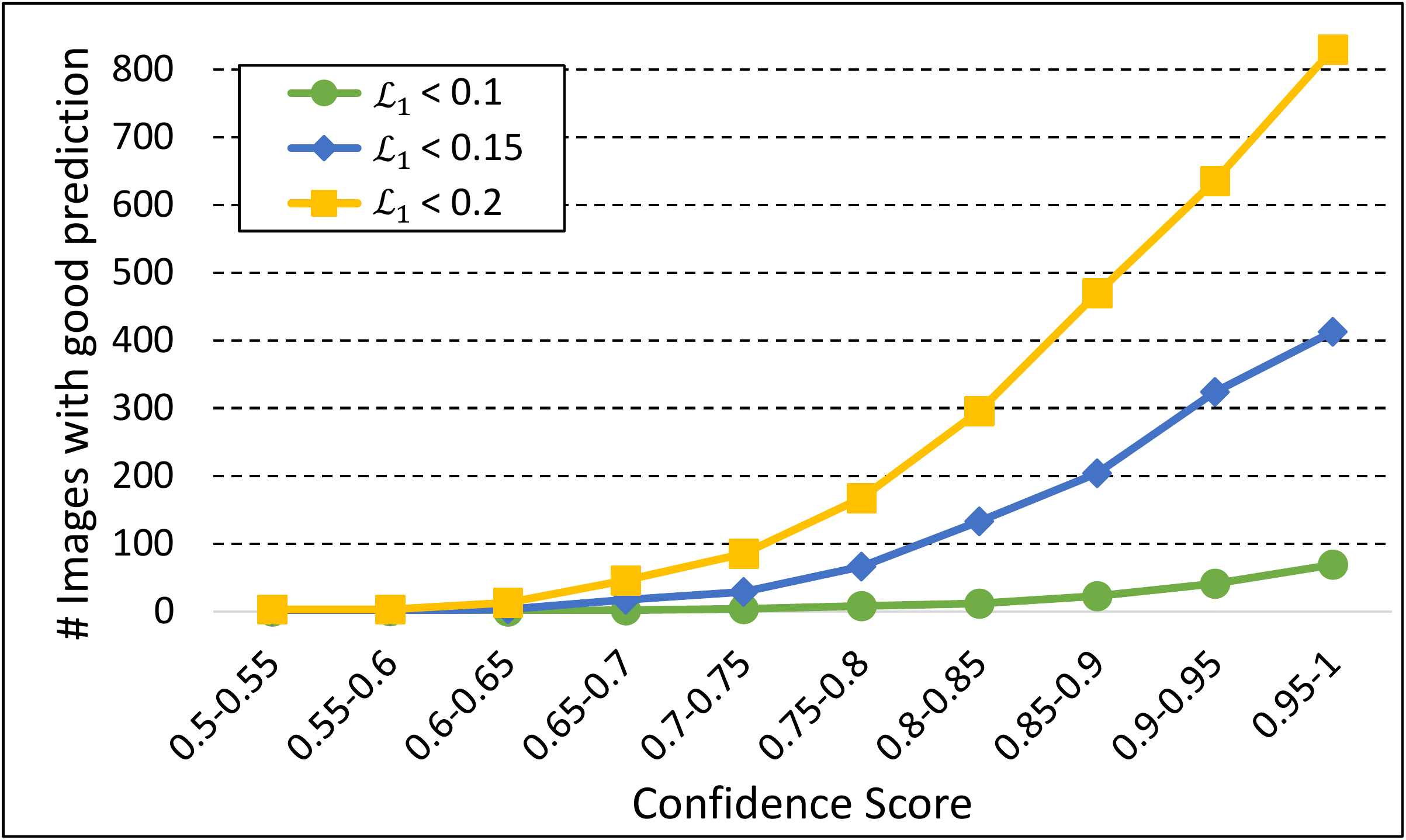}
  \end{center}
  \caption{\scriptsize{Correlation between the inpainting quality and confidence score.}}
  \label{fig:static_data}
     \end{minipage}
    \hspace{2mm}
    \begin{minipage}[c]{.48\linewidth}
\centering
	\footnotesize{
  		\begin{tabular}{ccccc}
			\includegraphics[width=0.2\columnwidth]{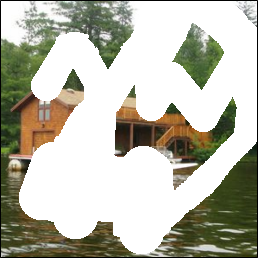} &
			\includegraphics[width=0.2\columnwidth]{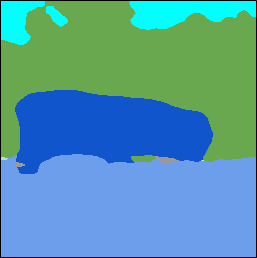} &
			\includegraphics[width=0.2\columnwidth]{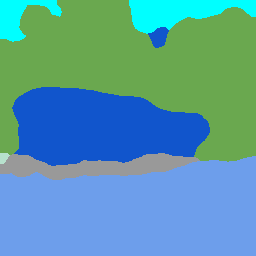} &
			\includegraphics[width=0.2\columnwidth]{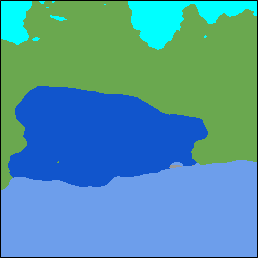} &
			\includegraphics[width=0.2\columnwidth]{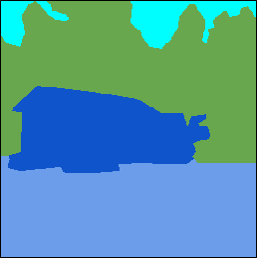} \\
			\includegraphics[width=0.2\columnwidth]{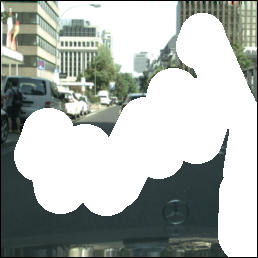} &
			\includegraphics[width=0.2\columnwidth]{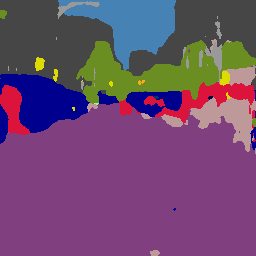} &
			\includegraphics[width=0.2\columnwidth]{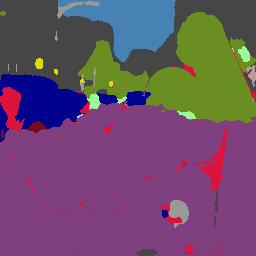} &
			\includegraphics[width=0.2\columnwidth]{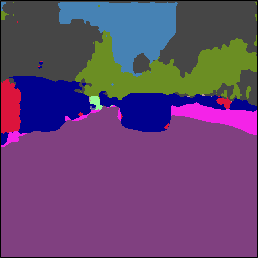} &
			\includegraphics[width=0.2\columnwidth]{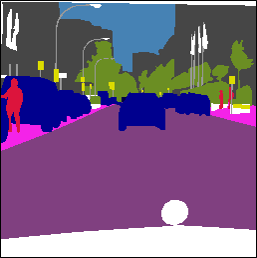}  \\
			\scriptsize{Input} & \scriptsize{EC} & \scriptsize{SPG} & \scriptsize{SGE-Net} & \scriptsize{GT}
		    \end{tabular}}
    \caption{\scriptsize{Visual comparison on semantic segmentation between SGE-Net and the segmentation-after-inpainting solutions. `EC' and `SPG' stand for EC+DPN/Deeplab and 
    SPG+DPN/Deeplab, respectively.}}
        \label{fig:segmentation-result}
    \end{minipage}
\end{figure}

We then further verify the effectiveness of the pixel-wise confidence scores by validating the correlation between the confidence scores and the $\mathcal{L}_1$ loss of the completed images with respect to the ground-truth that can be used to measure of the fidelity of inpainted pixels. We randomly select 9,000 images out of all the training and testing images from the two datasets with centering and irregular-hole settings, and calculate the average $\mathcal{L}_1$ loss and the confidence scores of all pixels in the missing region. As demonstrated in Fig.~\ref{fig:static_data}, the number of good-fidelity images with $\mathcal{L}_1$ loss increases with the segmentation confidence score (lower $\mathcal{L}_1$ means higher quality of the predicted image), implying the segmentation confidence score well serves the purpose of a metric of evaluating the accuracy of inpainted image content.
 
\textbf{Impact of Semantic Segmentation}
The success of semantics-guided inpainting largely relies on the quality of semantic segmentation map. Here we investigate the impact of segmentation accuracy on image inpainting. We conduct comparison between SGE-Net with segmentation maps generated by state-of-the-art segmentation tools and SGE-Net with human-labeled maps. We utilize the DPN model \cite{liu2017deep} pre-trained on \cite{wang2018recovering} as the segmentation tool for \textbf{Outdoor Scenes} as it is the only released model on the dataset. We select the Deeplab v3+ model \cite{chen2018encoder} for \textbf{Cityscapes} due to its superior performance on that dataset.

As shown in Table~\ref{tab:segmentation-source}, the performance degradation of our SGE-Net trained on imperfect semantic annotations is not significant, meaning that our model can still do a reasonably good job even trained on model-generated semantic annotation.  More subjective quality comparisons are provided in supplementary material. Note that the segmentation maps, either human-annotated or model-generated, are only used in the training stage of our model. While completing an image, SGE-Net itself can automatically generate the inpainted image and segmentation map simultaneously, without the need of the semantic annotations. 

\begin{wraptable}{r}{0.4\textwidth}
\centering
	    \linespread{1}
\caption{\scriptsize{Objective quality comparison on model trained by automatic segmentation (Auto-segs) and Human-labeled semantics (Label-segs).}}
    \setlength\tabcolsep{4pt}
    \resizebox{\linewidth}{!}{
    \begin{tabular}{lclc}
    \toprule
    
    \multicolumn{2}{c}{\scriptsize{Outdoor Scenes}} & \multicolumn{2}{c}{\scriptsize{Cityscapes}} \\
    \cmidrule(lr){1-2} \cmidrule(lr){3-4} 
    \scriptsize{Methods} & \scriptsize{PSNR} & \scriptsize{Methods} & \scriptsize{PSNR}   \\ \midrule
    \scriptsize{Auto-segs} & \scriptsize{20.19} & \scriptsize{Auto-segs} & \scriptsize{22.94} \\ 
    \scriptsize{Label-segs} & \bf{\scriptsize{20.53}} & \scriptsize{Label-segs}  & \bf{\scriptsize{23.41}}  \\ 

    \bottomrule
    \end{tabular}
    }    \label{tab:segmentation-source}
\end{wraptable}

We also conduct experiments to validate whether the iterative interplay between inpainting and semantic segmentation outperforms the traditional non-iterative segmentation-after-inpainting strategy in semantic segmentation. We compare the segmentation maps generated by SGE-Net itself with initial segmentation maps extracted from images completed by the baselines. As compared in Fig.~\ref{fig:segmentation-result}, the results show that SGE-Net evidently beats the segmentation-after-inpainting methods since SGE-Net leads to  more accurate semantic assignments and object boundaries, thanks to its joint-optimization of semantics and image contents. 

\section{Conclusion}
In this paper, a novel SGE-Net with semantic segmentation guided scheme was proposed to complete corrupted images of mixed semantic regions. To address the problem of unreliable semantic segmentation due to missing regions, we proposed a progressive multi-scale refinement mechanism to conduct interplay between semantic segmentation and image inpainting. Experimental results demonstrate that the mechanism can effectively refines poorly-inferred regions through segmentation confidence evaluation to generate promising semantic structures and texture details in a coarse-to-fine manner.

\section{Acknowledgement}
This work was supported in part by National Natural Science Foundation of China under Grant 91738302, 61671336, by Natural Science Foundation of Jiangsu Province under Grant BK20180234.

\clearpage
%
%
\bibliographystyle{splncs04}
\bibliography{egbib}
\end{document}